\documentclass[preprint,12pt]{elsarticle}

\usepackage{amssymb}
\usepackage{amsmath}
\usepackage{multirow}
\usepackage{multicol}
\usepackage{hyperref} 
\usepackage[T1]{fontenc} 

\usepackage[utf8]{inputenc}

\usepackage{lmodern} 
\usepackage[margin=2.2cm]{geometry}

\journal{Nuclear Physics B}

\begin{document}

\begin{frontmatter}



\title{MorphGen: Morphology-Guided Representation Learning for Robust Single-Domain Generalization in Histopathological Cancer Classification}


 \author[label1]{Hikmat Khan \corref{cor1}} \ead{hikmat.khan@osumc.edu}
 \author[label2]{Syed Farhan Alam Zaidi} \ead{farhanzaidi@cau.ac.kr}
 \author[label3]{Pir Masoom Shah}  \ead{pirmasoomshah@csu.edu.cn}
 \author[label4]{Kiruthika Balakrishnan} \ead{bkiruthi@uic.edu}
 \author[label5]{Rabia Khan}  \ead{rabiakhan@kust.edu.pk}
 \author[label6]{Muhammad Waqas}   \ead{mwaqas@mdanderson.org}
 \author[label6]{Jia Wu}  \ead{jwu11@mdanderson.org}

\cortext[cor1]{Corresponding author.}

\affiliation[label1]{organization={Department of Pathology, College of Medicine, The Ohio State University Wexner Medical Center},
            addressline={},
            city={Columbus},
           postcode={43210},
             state={OH},
             country={USA}}

 \affiliation[label2]{organization={Department of Computer Science and Engineering, Chung-Ang University},
             addressline={},
             city={Seoul},
             postcode={06974},
             state={},
             country={South Korea}}

 \affiliation[label3]{organization={School of Computer Science and Engineering, Central South University},
             addressline={},
             city={Changsha},
             postcode={410010},
             state={},
             country={China}}

 \affiliation[label4]{organization={College of Medicine Rockford, National Center for Rural Health Professions, University of Illinois},
             addressline={},
             city={Rockford},
             postcode={},
             state={IL},
             country={USA}}

 \affiliation[label5]{organization={Institute of Computing, Kohat University of Science and Technology},
             city={Kohat},
             state={Khyber Pakhtunkhwa (KPK)},
             country={Pakistan}}

 \affiliation[label6]{organization={Department of Imaging Physics, The University of Texas MD Anderson Cancer Center},
             addressline={},
             city={Houston},
             postcode={},
             state={TX},
             country={USA}}


\begin{abstract}
Domain generalization in computational histopathology is hindered by heterogeneity in whole slide images (WSIs), caused by variations in tissue preparation, staining, and imaging conditions across institutions. Unlike machine learning systems, pathologists rely on domain-invariant morphological cues—such as nuclear atypia (e.g., enlargement, irregular contours, hyperchromasia, chromatin texture, and spatial disorganization), structural atypia (abnormal architecture and gland formation), and overall morphological atypia—that remain diagnostic across different settings. Motivated by this, we hypothesize that explicitly modeling biologically robust nuclear morphology and spatial organization will enable the learning of robust cancer representations that are resilient to domain shifts. We propose MorphGen (Morphology-Guided Generalization), a novel method that integrates histopathology images, augmentations, and nuclear segmentation masks within a supervised contrastive learning framework. By aligning latent representations of images and nuclear masks, MorphGen prioritizes diagnostic features-nuclear and morphological atypia, and spatial organization—over staining artifacts and domain-specific features. To further enhance out-of-distribution (OOD) robustness, we incorporate stochastic weight averaging (SWA), steering optimization toward flatter minima. Attention map analyses revealed that MorphGen primarily relies on nuclear morphology, cellular composition, and spatial cell organization within tumors or normal regions for final classification. Finally, we demonstrate the resilience of the learned representations from our proposed method to image corruptions (e.g., staining artifacts) and adversarial attacks, thus showcasing not only out-of-domain generalization, but also addressing critical vulnerabilities in current deep learning systems for digital pathology. The code, datasets, and trained models are publicly available at \href{https://github.com/hikmatkhan/MorphGen/}{https://github.com/hikmatkhan/MorphGen}.
\end{abstract}

\begin{graphicalabstract}
\includegraphics[width=0.95\textwidth]{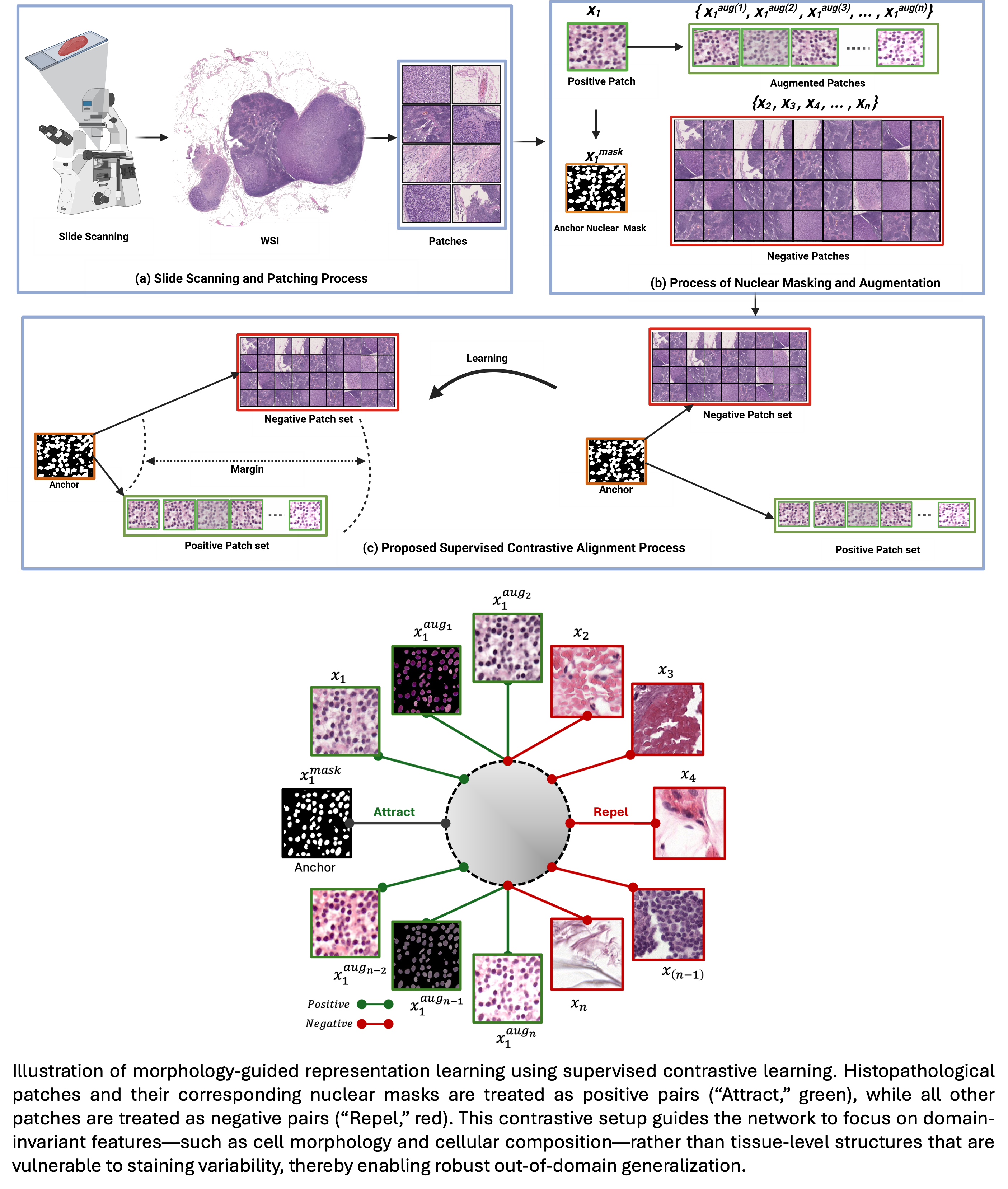}

\end{graphicalabstract}

\begin{highlights}

\item Proposed MorphGen achieves high accuracy in histopathology-based cancer classification

\item Employs morphology-guided representation learning to learn robust, domain-invariant features.

\item Mitigates domain shift and achieves consistent cross-domain performance across multiple public datasets.

\item Maintains strong accuracy under diverse staining variations, natural noise, and adversarial perturbations.

\item Enhances out-of-domain generalizability of deep learning, enabling reliable deployment in real-world clinical pathology applications.

\end{highlights}


\end{frontmatter}




\section{Introduction}
Computational Pathology (CPath) is considered the gold standard for microscopic examination of tissue structure and cellular morphology \cite{aggarwal2025artificial}, establishing itself as essential for accurate cancer diagnosis and informed therapeutic decision-making \cite{mcgenity2024artificial, datwani2025artificial}. Additionally, the integration of artificial intelligence, particularly Deep Learning (DL), has significantly advanced several clinical applications, including survival prediction \cite{ tsai2023histopathology,  javanmard2025artificial, piedimonte2025predicting, khan2020cascading}, gene mutation and expression profiling \cite{jahani2025efficacy, jung2025h}, treatment response prediction\cite{khan2025predicting},  tumor subtyping \cite{koziarski2024diagset, lomenie2022can}, biomarker discovery \cite{shulman2025path2space, shah2025deepdtagen, campanella2025real}, and tumor microenvironment (TME) analysis \cite{shao2024tumor, xie2024artificial}, among others \cite{muneer2025classical, mcgenity2024artificial,balakrishnan2025artificial, waqas2024exploring,hussain2025alzformer, datwani2025artificial}.

Despite these advancements, the clinical deployment of deep learning models remains limited due to the persistent challenge of domain shift \cite{9847099, jahanifar2025domain}, where non-biological variations arise from deviations in acquisition systems, such as firmware updates, altering illumination spectra or technician-driven recalibration of the camera, introducing subtle yet impactful perturbations to the joint distribution of color, texture, and noise characteristics \cite{pocevivciute2023detecting}. Multi-center benchmark studies have reported a consequent performance degradation \cite{ochi2024registered}. Moreover, building large-scale, diverse, multi-institutional cohorts is logistically challenging \cite{ochi2024registered}. Critically, this domain shift is systematic yet label-agnostic, causing decision boundaries to distort unevenly across classes, thereby making the model less effective in capturing these variations. These challenges underscore the critical need to prioritize domain adaptation as a core design principle in model training, which ensures robust generalization on out-of-distribution (OOD) across diverse imaging protocols \cite{pocevivciute2023detecting,carretero2024enhancing}, while facilitating seamless integration into clinical workflows to enhance patient outcomes \cite{jahanifar2025domain, tomar2024nuclear}.

Current approaches to address domain shifts can be grouped into four categories: (1) Stain normalization techniques (e.g., Macenko \cite{macenko2009method} and RandStainNA \cite{shen2022randstainna}) standardize color but frequently alter diagnostically critical features; (2) Data augmentation strategies (e.g., Learning to Diversify (L2D) \cite{wang2021learning}), create synthetic variations but risk generating unrealistic artifacts; (3) Feature-space methods (e.g., Representation Self-Challenging (RSC) \cite{huang2020self} ), that eliminate unstable features without biological grounding; and (4) Single-domain generalization approaches, which train on single-institution data. For example, RSC \cite{huang2020self}  discards high-gradient features, promoting stable feature reliance. Adversarial Domain Augmentation (ADA) \cite{volpi2018generalizing} uses adversarial examples to enrich the source domain.  Tomar et al.\cite{tomar2024nuclear} proposed aligning image and mask representations via Euclidean distance-based regularization to encourage robust feature learning. However, current methods largely overlook biological invariants—particularly nuclear morphology, tissue architecture, and cellular organization—that pathologists rely on for consistent diagnosis across clinical settings (see Section \ref{sec:motivation} for more details).

In this paper, we hypothesize that explicitly modeling domain-specific, invariant morphological characteristics, such as cellular organization, nuclear shape, and size — central to cancer diagnosis yet resilient to technical variability- will enable models to learn biologically relevant and generalizable representations, leading to consistent diagnostic performance across heterogeneous clinical settings. 

\begin{figure*}[htbp!]
    \centering
    \includegraphics[width=\textwidth]{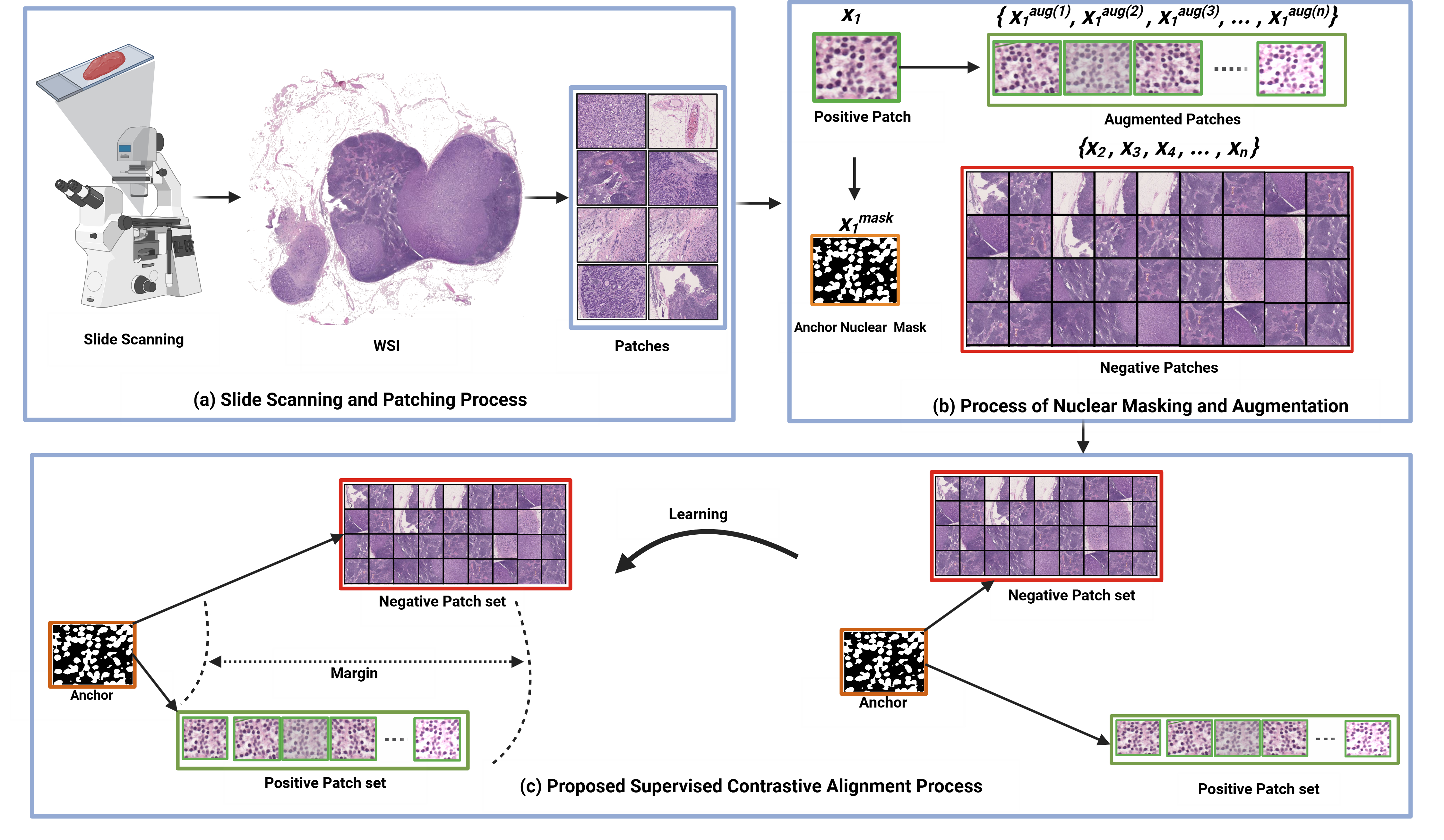}
    \caption{Illustration of Morphology-guided Representation Alignment for Cancer Classification, utilizing supervised contrastive learning to align histopathological images and nuclear masks in a shared latent space, with "Attract" (green) and "Repel" (red) relationships guiding the feature representation learning.}
    \label{fig:abstract}
\end{figure*}

To this end, we propose MorphGen, a 
\textbf{Morph}ology-\textbf{G}uided G\textbf{en}eralization approach for robust domain generalization of Histopathological Cancer Classification, with a schematic overview illustrated in Figure~\ref{fig:abstract}. MorphGen incorporates histopathology images, their augmentations, and nuclear segmentation masks during training through supervised contrastive learning to enforce latent space alignment between images and their corresponding nuclear masks, thereby encouraging the model to learn diagnostically relevant features (such as nuclear atypia, cellular organization, and nuclear morphology) while suppressing domain-specific artifacts \cite{khosla2020supervised}. Additionally, to further improve model robustness and convergence, we apply optimization techniques such as stochastic weight averaging (SWA) \cite{izmailov2018averaging, stoica2023zipit, wortsman2022model}, guiding the model toward flatter loss minima \cite{arpit2022ensemble, cha2021swad, rame2022diverse, izmailov2018averaging}, and thereby enhancing generalization across diverse domains \cite{jain2023dart, li2022simple, shu2023clipood, wang2023sharpness}. Experimental results across three datasets—CAMELYON17, BCSS, and OCELOT—demonstrate that MorphGen learns robust and biologically meaningful representations, achieving consistently improved out-of-domain accuracy over strong baselines. The model not only generalizes across datasets but also across organ types in cancer classification. Furthermore, ablation studies reveal that the learned representations exhibit resilience against both natural corruptions and adversarial perturbations, highlighting the strength of morphology-guided representation learning for out-of-domain generalization.

The proposed MorphGen offers the following key contributions:
\begin{itemize}
    \item We introduce a novel representation alignment framework that jointly leverages whole-slide images, their augmented variants, and nuclear segmentation masks through supervised contrastive learning. This approach encourages the model to align latent representations of morphological features corresponding to nuclear structures, thereby fostering the learning of biologically meaningful and domain-invariant features.
    \item We incorporate optimization strategies (such as stochastic weight averaging) to guide training toward flatter and wider loss minima, thereby enhancing out-of-domain robustness. This effectively mitigates the adverse impact of domain shifts caused by technical variability across datasets from diverse clinical institutions, resulting in further improvements in out-of-domain generalization.
    \item We provide extensive empirical validation to demonstrate that the robustness of the learned representations in MorphGen surpasses that of baseline methods for robust cancer classification in the presence of both natural (e.g., staining artifacts) and adversarial perturbations. This underscores the importance of morphology-guided representation learning and its strong potential for reliable deployment in real-world clinical settings. 
\end{itemize}
The proposed MorphGen consistently outperforms state-of-the-art domain adaptation and generalization techniques, demonstrating superior diagnostic accuracy on unseen institutions. Additionally, it exhibits enhanced resilience against both natural and adversarial perturbations, underscoring its strong potential for reliable deployment in real-world clinical settings.

\section{Biological Motivation Beyond MorphGen for Robust Out-of-Domain Generalization}
\label{sec:motivation}

Incorporating nuclear morphological atypia as a form of domain knowledge is critical for guiding deep learning models toward clinically meaningful and domain-invariant representations, thereby enhancing classification accuracy and generalization across heterogeneous patient populations and clinical sites. This imperative is particularly pronounced in digital pathology, where expert interpretation of nuclear morphology—such as size, shape, and chromatin texture—is central to diagnosis but often neglected by end-to-end, data-driven architectures. Morphological atypia, encompassing both structural irregularities and nuclear abnormalities, remains foundational to cancer classification, tumor grading, survival prediction, and therapeutic response modeling. In recognition of its diagnostic utility, several computational approaches have sought to explicitly integrate morphological priors into deep learning workflows. For instance, Shen et al. \cite{shen2023development} fused convolutional neural networks with traditional models trained on fine-grained nuclear descriptors to predict neoadjuvant chemotherapy (NAC) response from biopsy images. Similarly, David et al. \cite{dodington2021analysis} demonstrated that AI-based analysis of nuclear intensity and textural heterogeneity significantly enhances response prediction in core biopsies. Extending this paradigm, Wang et al. \cite{wang2024nuclei} introduced NPKC-MIL, a constrained multiple instance learning framework that integrates (1) transfer learning-based patch features with attention aggregation, (2) nuclei graphs built on K-NN topology enriched with handcrafted geometric and textural descriptors, and (3) multi-scale feature fusion to refine slide-level classification. Their model not only surpasses conventional MIL baselines in accuracy and interpretability but, through analysis of nearly 1 million nuclei, reveals systematic differences in cancerous versus normal nuclei in terms of size, shape, entropy, and spatial disorganization—underscoring the clinical relevance of morphological atypia.

The diagnostic significance of nuclear morphology is further supported by foundational studies across cancer types. Aberrations in nuclear shape and organization, as established by Neil et al. \cite{carleton2018advances} and others, serve as hallmarks of cellular dysregulation and are routinely leveraged as biomarkers in breast \cite{chen2015new, nawaz2016computational}, prostate \cite{carleton2018advances}, and other epithelial cancers \cite{fischer2020nuclear}. Comparative histopathological studies reveal substantial differences in tissue architecture and cellular composition between normal and diseased states \cite{melssen2023barriers}, with immune cell spatial organization emerging as a key correlate of inflammation, disease progression, and survival \cite{sherwood2004mechanisms, melssen2023barriers}. 

In particular, apoptotic cells—crucial for tissue homeostasis and the elimination of damaged or cancerous cells—exhibit distinctive morphological features such as cell shrinkage, chromatin condensation, and nuclear fragmentation, which may serve as informative latent variables in deep models, providing biologically grounded signals that remain consistent across domains \cite{park2023diversity}. This growing appreciation for structural biomarkers has catalyzed the development of high-quality nuclei segmentation datasets \cite{gamper2020pannuke, graham2019mild, graham2021lizard, graham2019hover, grossman2016toward, kumar2017dataset, mahbod2021cryonuseg, naylor2017nuclei, naylor2018segmentation, sirinukunwattana2017gland, verma2021monusac2020, vu2019methods}, along with advanced segmentation models \cite{graham2019hover, han2023ensemble, hollandi2022nucleus, horst2024cellvit, johnson2020automatic, schmidt2018cell, stringer2021cellpose, horst2025cellvit, yuan2025crns}, which now underpin critical downstream tasks such as mitotic count estimation and immune cell identification. Despite these advances, much of the current work remains confined to in-domain classification. Notably, DeepCMorph \cite{ignatov2024histopathological} introduced a two-stage pipeline where nuclei segmentation and cell-type annotation maps are integrated into classification workflows. This morphology-aware model achieved over 4\% improvement in accuracy compared to large transformer-based models pretrained on millions of histology images \cite{wang2022transformer}, and demonstrated superior generalizability on small-scale microscopy datasets. Beyond individual cell features, colorectal carcinoma (CRC) exemplifies the clinical importance of tissue-level morphological heterogeneity \cite{pai2022quantitative}. CRC exhibits diverse architectural patterns—including mucinous morphology, tumor budding, stromal desmoplasia, and poorly differentiated clusters—that are associated with distinct prognoses and underlying molecular alterations (e.g., mismatch repair deficiency) \cite{ueno2013objective, idos2020prognostic, lee2020analysis}. These morphological attributes, which include both cellular and spatial features such as immune infiltration patterns and tumor-stroma ratios, are difficult to quantify manually but offer rich supervisory signals for deep learning.  There is a pressing need for computational tools capable of not only quantifying these features but also learning biologically grounded representations to inform therapy prediction, aid in early diagnosis and prognosis, support the development of personalized therapeutic strategies, and be adaptable for deployment in diverse clinical settings.

Building on these insights, we posit that three core pillars—(1) morphological atypia, (2) spatial tissue architecture, and (3) domain-informed feature learning—are essential for robust generalization in computational pathology. Motivated by the consistency of nuclear morphology across diverse staining and imaging conditions, we propose a novel framework that integrates segmentation-derived priors into a supervised contrastive learning paradigm. To our knowledge, this is the first method to directly embed segmentation masks into representation learning, thereby steering the model toward biologically grounded, domain-stable features. Specifically, we implement a representation alignment strategy that simultaneously leverages histopathology images, their augmentations, and corresponding nuclear masks to enforce feature consistency across views. Supervised contrastive loss is employed to shape the latent space, prioritizing features that align with expert morphological cues. In addition, we incorporate stochastic weight averaging (SWA) to encourage convergence to flatter minima, improving robustness. This unified approach not only enhances out-of-distribution (OOD) classification performance but also improves resilience to staining artifacts and adversarial perturbations—two critical failure points for current deep learning models in clinical pathology. The comparative capabilities of the MorphGen method are summarized in Table \ref{tab:domain_shift_methods}.

\begin{table*}[t]
\footnotesize
\scriptsize
\centering
\caption{Comparison of Domain Shift Mitigation Methods in WSI Analysis}
\label{tab:domain_shift_methods}
\renewcommand{\arraystretch}{1.3}
\setlength{\tabcolsep}{4pt}
\begin{tabular}{
p{3.5cm} 
p{1.6cm} 
p{1.9cm} 
p{1.6cm} 
p{2.1cm} 
p{4.5cm} 
}
\hline
\textbf{Method} & 
\textbf{Stain Variability} & 
\textbf{Resolution Discrepancies} & 
\textbf{Feature Invariance} & 
\textbf{Diagnostic Feature Preservation} & 
\textbf{Limitations} \\
\hline
Macenko \cite{macenko2009method} 
& Yes
& No
& No
& No
& Alters diagnostic features; no resolution handling \\
RandStainNA \cite{shen2022randstainna} 
& Yes
& Yes
& No
& No
& Risks unrealistic artifacts \\
L2D \cite{wang2021learning} 
& No
& Yes
& Yes
& No
& Struggles with morphological variability \\
RSC \cite{huang2020self}
& No
& No
& Yes
& No
& Risks loss of discriminative features \\
Proposed (MorphGen) 
& Yes
& Yes
& Yes
& Yes
& Requires careful tuning of SWA schedules \\
\hline
\end{tabular}
\end{table*}


\section{Related Work}
\subsection{Stain Normalization Based Methods}
Traditional stain normalization methods aim to standardize the color appearance of hematoxylin and eosin (H\&E)-stained histopathology images by transforming source images to match the color distribution of a target. Early approaches, such as Macenko normalization \cite{macenko2009method}, employed stain vector estimation to perform this transformation. Vahadane et al. \cite{vahadane2016structure} improved upon this by decoupling stain density maps from color appearance, enabling the more flexible recombination of histological features. Reinhard et al. \cite{reinhard2001color} introduced a technique that adjusts global color statistics in an alternative color space, effectively transferring the color characteristics of the target image to the source. More recently, RandStainNA \cite{shen2022randstainna} proposed a data-driven augmentation strategy, generating virtual templates from Gaussian distributions of training image statistics in LAB space, rather than using fixed references. While these methods are effective at reducing inter-institutional color variability (as comprehensively reviewed in \cite{hoque2024stain}), they fundamentally operate on low-level color transformations. Furthermore, color augmentation—applying random perturbations to hue, saturation, and brightness—can improve model robustness; however, the optimal degree of perturbation is challenging to tune, as excessive changes may introduce unnatural color distributions that harm model performance \cite{ochi2024registered}. This can inadvertently distort diagnostically critical features, such as nuclear morphology and tissue architecture. In summary, stain normalization techniques primarily address color heterogeneity, often overlooking higher-order biological structures vital for accurate diagnosis. In contrast, our approach shifts the focus from color manipulation to biologically grounded features, explicitly targeting morphological atypia, nuclear shape, and cellular organization (see Section \ref{sec:methodology})

\subsection{Data Augmentation Based Methods}
Data augmentation has emerged as a critical strategy for improving domain generalization in histopathology image analysis. Tellez et al. \cite{tellez2019quantifying} conducted a comprehensive evaluation, demonstrating that stain color augmentation significantly enhances model performance on external test sets compared to stain normalization alone. Subsequent work has focused on developing domain-specific augmentation techniques tailored to histopathology: Faryna et al. \cite{faryna2021tailoring} adapted RandAugment \cite{cubuk2021practical} by incorporating pathology-relevant transformations while eliminating unrealistic image modifications. Specialized methods, such as Tellez et al.’s H\&E-specific augmentation \cite{tellez2018h}, have shown particular effectiveness for tasks like mitosis detection, while Pohjonen et al.’s StrongAugment \cite{pohjonen2022augment} employs variable-length transformation sequences to boost generalization. More sophisticated approaches, such as Marini et al \cite{marini2023data} Data-Driven Colour Augmentation (DDCA), introduce quality control by assessing augmented images against a reference database. Notably, Faryna et al. \cite{faryna2021tailoring} found that both manual and automated augmentation strategies can achieve competitive performance, suggesting that careful design and selection of augmentations may be as crucial as their algorithmic complexity. While these approaches improve model robustness, they primarily address color and texture variations without explicitly preserving or enhancing diagnostically critical features such as nuclear morphology, tissue architecture, and cellular organization—a limitation our work directly addresses through biologically grounded representation learning.

\subsection{Single-Domain Generalization (SDG) Based Method} 
Recent advances in single-domain generalization (SDG) have led to the development of innovative approaches to enhance model robustness using only single-source training data. Representation Self-Challenging (RSC) \cite{huang2020self} operates by selectively discarding high-gradient activation features during training, thereby forcing the model to rely on more stable representations. Adversarial methods such as ADA \cite{volpi2018generalizing} employ iterative generation of adversarial examples to augment the source domain, while its extension M-ADA \cite{qiao2020learning} combines this approach with Wasserstein Auto-Encoders \cite{tolstikhin2017wasserstein} and meta-learning for enhanced sample generation. Alternative strategies include the Progressive Domain Expansion Network (PDEN) \cite{li2021progressive}, which utilizes multiple autoencoders to progressively expand the training distribution, and L2D's learnable style-complement module \cite{wang2021learning}, which maximizes image diversity while preserving diagnostic semantics through adversarial training. Tomar et al. \cite{tomar2024nuclear} proposed aligning image and mask representations using Euclidean distance-based regularization, a method closely related to ours. However, euclidean distance has been shown to produce less robust and generalizable features compared to contrastive learning frameworks \cite{9226466}. While these methods demonstrate promising results in handling technical variations, they fundamentally overlook the biological invariants—particularly nuclear morphology and tissue architecture—that pathologists inherently rely on for consistent diagnosis across different clinical settings. This critical limitation motivates our biologically grounded approach that explicitly models these diagnostic features while maintaining robustness to domain shifts.

\subsection{Domain Adaptation Based Methods}
Unlike single-domain generalization, domain adaptation methods require access to target domain samples during training. In histopathology, these approaches predominantly leverage generative adversarial networks (GANs) and their variants \cite{goodfellow2014generative}. StainGAN \cite{shaban2019staingan} employs CycleGAN \cite{zhu2017unpaired} to transform source domain images into the target domain's appearance space, while Residual CycleGAN \cite{zhu2017unpaired} modifies this approach by having the generator predict inter-domain residuals rather than complete images. In \cite{zhou2019enhanced}, the incorporation of stain color matrices as auxiliary generator inputs was further refined to improve training stability. \text{NST\_AD\_HRNet} \cite{nishar2020histopathological} combines neural style transfer \cite{gatys2016image, johnson2016perceptual} with GANs to preserve source image content while adopting target domain style characteristics. Earlier approaches \cite{yuan2018neural, cho2017neural} introduced a two-stage process involving grayscale conversion followed by target-guided recoloring. While these methods demonstrate technical innovation, domain adaptation remains fundamentally impractical for many clinical applications due to its dependence on target domain data, a limitation exacerbated by recent evidence suggesting it may yield poorer generalization than stain normalization and color augmentation approaches \cite{stacke2020measuring}. This underscores the clinical relevance of our single-domain generalization framework, which achieves robust performance without requiring access to data from target institutions.

\section{Methodology}
\label{sec:methodology}
\begin{figure*}[h]
    \centering
    \includegraphics[width=0.80\textwidth]{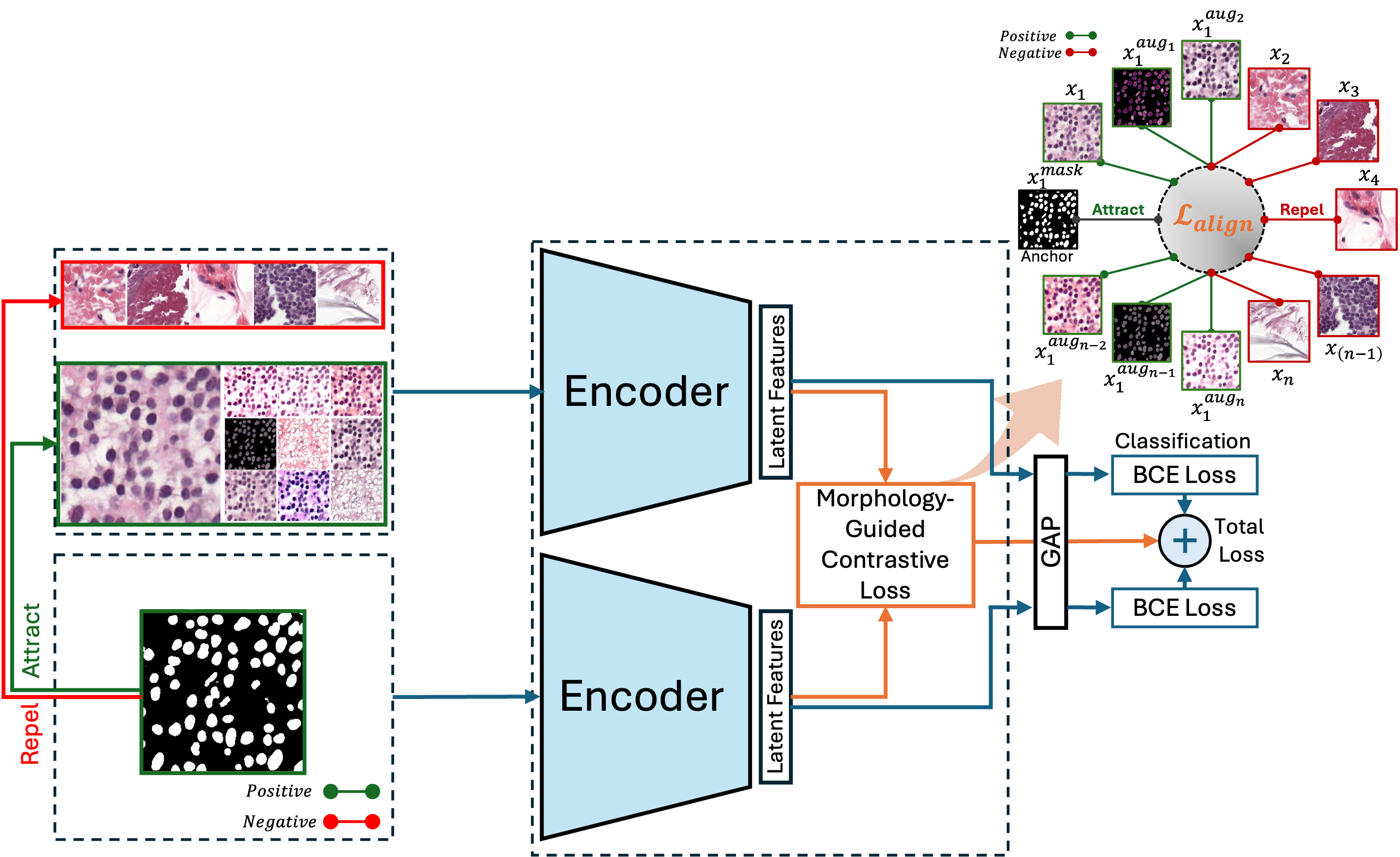}
    \caption{Architecture of  MorphGen (\textbf{Morph}ology-\textbf{G}uided g\textbf{en}eralization in histopathological
cancer classification), illustrating the use of a shared ResNet-based encoder to process histopathology patches and nuclear masks, with a composite loss function combining Morphology-Guided Contrastive Loss and Binary Cross-Entropy (BCE) Loss for robust feature extraction and classification.}
    \label{fig:main}
\end{figure*}

In this section, we present the details of the proposed MorphGen (Morphology-guided Generalization) framework for histopathological cancer classification. The overall architecture is illustrated in Figure~\ref{fig:main}.

\subsection{Shared Encoder}

MorphGen's encoder is built upon a ResNet-based architecture (initialized with ImageNet weights), which serves as the shared backbone for processing histopathology patches, associated augmentations, and segmentation masks, thereby facilitating robust feature extraction, illustrated in Figure~\ref{fig:main}. The encoder extracts latent representations from each input pair and projects them into a unified embedding space.
By sharing the encoder across these inputs, the model optimizes computational efficiency while maintaining consistent feature embeddings, which are crucial for computational pathology workflows.
This weight-sharing architecture encourages the network to concentrate on nuclear morphology during training and maintain consistent feature embeddings to tackle OOD generalization while eliminating the need for segmentation masks during inference, thus ensuring compatibility with real-world diagnostic scenarios. By sharing the encoder across these inputs, the model optimizes computational efficiency while maintaining consistent feature embeddings, which are crucial in digital pathology workflows and ensuring compatibility with real-world diagnostic scenarios.
\subsection{Objective Function}
In this section, we provide a detailed discussion of the objective function.
\subsubsection{Morphology-Guided Supervised Contrastive Loss}

MorphGen employs a triplet-based supervised contrastive learning framework, where the nuclear segmentation mask serves as the anchor to learn discriminative and relevant features of nuclear morphology and spatial cellular organization. Intuitively, the contrastive loss minimizes the distance between the embedding of the anchor (i.e., nuclear segmentation mask) and positive patches (i.e., corresponding patches or their augmented versions), while maximizing the distance to negative patches (i.e., other patches in the batch) within a shared latent space. This objective promotes the learning of features rooted in nuclear morphology and spatial organization—structural cues that remain consistent across diverse staining protocols and imaging platforms.

Formally, given a batch of $N$ paired inputs, the loss is defined as:

\begin{align}
\mathcal{L}_{\text{align}}(\mathbf{z}^{\text{mask}}, \mathbf{z}^{\text{aug}}, \mathbf{z}) 
&= \mathcal{L}_{\text{attract}}^{+} + \mathcal{L}_{\text{repel}}^{-}
\end{align}
where,

\begin{align*}
    \mathcal{L}_{\text{attract}}^{+} = \frac{1}{N} \sum_{i=1}^N \left(1 - 
    \frac{ \mathbf{z}_i^{\text{mask}} \cdot \mathbf{z}_i^{\text{aug}} }
         { \left\| \mathbf{z}_i^{\text{mask}} \right\|_2 
           \left\| \mathbf{z}_i^{\text{aug}} \right\|_2 } \right)
\end{align*}

\begin{align*}
    \mathcal{L}_{\text{\text{repel}}}^{-} &= \frac{\lambda}{N(N-1)} \sum_{i=1}^N \sum_{\substack{j=1 \\ j \ne i}}^N 
    \max\left( 0, 
    \frac{ \mathbf{z}_j \cdot \mathbf{z}_i^{\text{mask}} }
    { \left\| \mathbf{z}_j \right\|_2 \left\| \mathbf{z}_i^{\text{mask}} \right\|_2 } 
    - \eta \right)^2
\end{align*}


\noindent
Here, $\mathbf{z}^{\text{mask}}$ and $\mathbf{z}^{\text{aug}}$ represent the embeddings of the nuclear masks and their corresponding histology patches (or augmentations), which are treated as positive pairs. Meanwhile, $\mathbf{z}$ signifies the embeddings of other histology patches within the batch, designated as negatives.  The hyperparameter $\eta$ serves as a similarity margin, imposing an upper limit on the similarity between negative pairs. The first term aims to minimize the distance between the embeddings of morphology-aligned inputs, ensuring that the model captures features essential to nuclear size, contour, chromatin texture, and local spatial patterns. The second term penalizes excessive similarity among mismatched pairs, thereby enhancing differentiability in the learned feature space. This contrastive formulation motivates the model to acquire morphology-aware, semantically rich features that improve generalization across various datasets, staining conditions, and imaging artifacts.

\subsubsection{Classification loss}
To ensure that the learned representations are not only morphologically aligned but also diagnostically significant, we incorporate the Morphology-Guided Contrastive Loss alongside a Binary Cross-Entropy (BCE) classification loss. The BCE loss is calculated for both the primary prediction $\hat{y}$ and a secondary prediction $\hat{y}'$, which is derived from the respective mask label and serves as a regularization signal during training. It is defined as:

\begin{equation}
\mathcal{L}_{\text{bce}} (y, \hat{y}, \hat{y}')  = \text{BCE}(y, \hat{y}) + \text{BCE}(y, \hat{y}').
\end{equation}

Here, the first term, $\text{BCE}(y, \hat{y})$, corresponds to patch-level classification, and the second term, $\text{BCE}(y, \hat{y}')$, corresponds to classification based on the mask-guided representation. Each term is computed as:

\begin{equation}
\text{BCE}(y, \hat{y}) = -\frac{1}{N} \sum_{i=1}^{N} \left[ y_i \log(\hat{y}_i) + (1 - y_i) \log(1 - \hat{y}_i) \right]
\end{equation}

\subsubsection{Total Loss}
The total loss integrates the contrastive alignment and classification losses as below.

\begin{equation}
\mathcal{L}_{\text{total}} = \mathcal{L}_{\text{align}} + \mathcal{L}_{\text{bce}}
\end{equation}

The proposed loss function addresses two primary challenges  in OOD : 1) it promotes the learning of domain-invariant features embedded in nuclear morphology and cellular organization, these structural cues remain consistent across various staining protocols and imaging platforms;  2) facilitating the extraction of diagnostic patterns that generalize robustly across different institutions and out-of-distribution datasets, and enahances the generalization capabilities of the model across various clinical settings and acquisition protocols.

\section{Experimental Setup}

\subsection{Datasets}

We utilize three publicly available histopathology datasets: the CAncer MEtastases in LYmph nOdes challeNge (CAMELYON17)\cite{litjens20181399}, the Breast Cancer Semantic Segmentation (BCSS), and the Overlapped Cell on Tissue (OCELOT). Each dataset provides diverse, high-resolution whole-slide images (WSIs) with expert-verified annotations. The CAMELYON17 dataset includes 1,000 hematoxylin and eosin (H\&E)-stained WSIs of lymph node sections with breast cancer metastases, sourced from five Dutch medical institutions. Each institution contributes 200 WSIs, with 10 slides per institution annotated at the pixel level for tumor regions, resulting in a total of 50 annotated slides. All images were digitized at 40× magnification. While institutions 0, 3, and 4 used the same scanner, institutions 1 and 2 utilized different devices. To enable a robust evaluation of domain generalization, we treat each institution as a distinct domain, capturing real-world variation in image acquisition pipelines. The BCSS dataset comprises 151 H\&E-stained whole-slide images (WSIs) of histologically confirmed triple-negative primary breast cancers, sourced from The Cancer Genome Atlas (TCGA) \cite{amgad2019structured}. All slides were digitized at 40× magnification. The OCELOT dataset comprises 303 H\&E-stained whole-slide images (WSIs) from The Cancer Genome Atlas (TCGA), covering six organ sites: bladder, endometrium, head and neck, kidney, prostate, and stomach \cite{ryu2023ocelot}. Two slides scanned only at 20× magnification were excluded, yielding 301 high-resolution slides. Together, these datasets enable a comprehensive assessment of model generalization across diverse tissue types and institutional imaging pipelines.

\subsection{Data Augmentation}
Data heterogeneity across institutions introduces batch effects stemming from differences in tissue processing, staining protocols, scanning equipment, and image post-processing. These embedded signatures can bias models towards learning spurious, site-specific representations, resulting in strong performance in the institution data but poor generalization to out-of-institution domains \cite{howard2021impact, fang2023sqad}. Recent studies have demonstrated that aggressive data augmentation effectively mitigates such batch effects by encouraging models to learn domain-invariant features across institutions \cite{fang2023sqad}. Following this approach, we apply aggressive data augmentation to histopathological patches to promote out-of-domain generalization. The augmentations include: random rescaling (0–20\%), aspect ratio distortion (0–10\%), rotation (0–360°), and adjustments to brightness (0–50\%), hue (0–10\%), contrast (0–70\%), and saturation (0–30\%), as well as Gaussian noise injection. As shown in Figures \ref{fig:abstract} and \ref{fig:main}, these transformations disrupt color, texture, and scale-related signals, increasing training difficulty, reducing overfitting, and encouraging the model to focus on learning domain-invariant features (e.g., nuclear morphology). This ultimately reduces inter-institute variability and enhances generalization performance.

\subsection{Baseline Comparisons}
We conducted a comprehensive evaluation of our proposed method against several representative baselines for single-domain generalization, including Macenko \cite{macenko2009method}, HoVerNet \cite{graham2019hover}, RandStainNA \cite{shen2022randstainna}, RSC \cite{huang2020self}, L2D \cite{wang2021learning}, SFL \cite{tomar2024nuclear}, ERM \cite{gulrajani2020search}, and DDCA \cite{marini2023data}. These baselines encompass a broad spectrum of strategies, including stain normalization (Macenko, RandStainNA), data augmentation (L2D), feature-space regularization (RSC, DDCA), and representation alignment (SFL). This selection ensures a thorough and balanced comparison against state-of-the-art single-domain generalization methods in histopathology.

\begin{figure*}[t]
    \centering
    \includegraphics[width=0.9999999\textwidth]{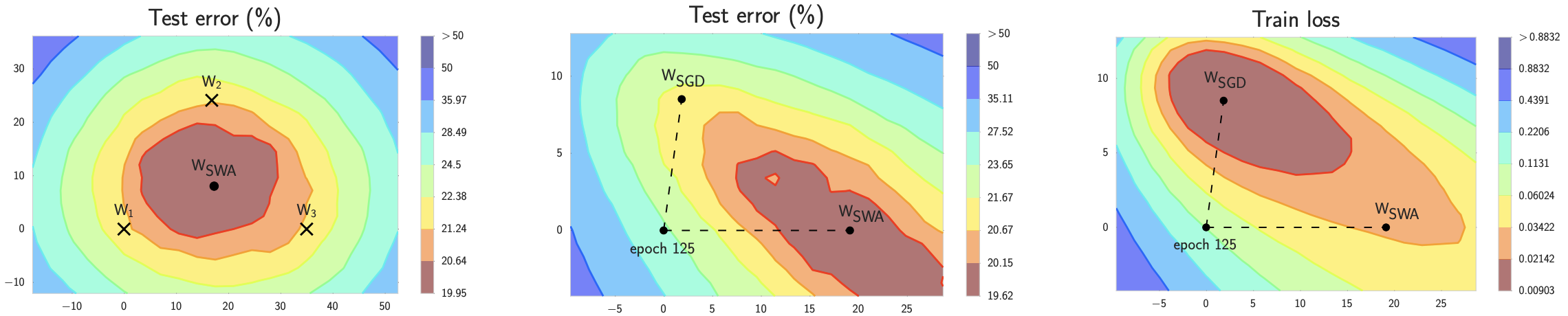}
    \caption{Illustration of stochastic weight averaging (SWA). \textbf{Left:} SWA produces a solution ($w_{\text{SWA}}$) lying in a flatter region of the test error surface compared to individual model weights ($w_1, w_2, w_3$). Such flatter regions (often called valleys of near-constant loss) encourage robustness to domain shifts compared to convergence to saddle points (where small perturbations or shifts in the domain can cause drastic changes in the loss surface), thereby enhancing out-of-domain generalization. \textbf{Middle:} Trajectory of standard stochastic gradient descent (SGD) ($w_{\text{SGD}}$) and the corresponding SWA solution on the test error surface. \textbf{Right:} Train loss surface illustrating how SWA finds a flatter minimum than standard SGD. (Figure borrowed from \cite{izmailov2018averaging}.)}
    \label{fig:swa}
\end{figure*}

\subsection{Training and Evaluation Details}
We conduct our training on the CAMELYON17 dataset \cite{litjens20181399}, treating each of the five Dutch medical centers as a separate domain, and evaluate performance on the CAMELYON17 test set. To assess out-of-domain generalization, models trained on CAMELYON17 are tested on the BCSS and OCELOT datasets, which are exclusively reserved for final evaluation to provide an unbiased measure of generalization across institutions. During training, patches of size 512x512 pixels are extracted at 40× magnification using the SFL codebase \cite{tomar2024nuclear}. We implement patient-level splits, reserving approximately 20–25\% of tiles per domain for validation. Our learning rate schedule includes a warm-up phase over the first 40 epochs, during which the rate is gradually increased from 0 to $1.5 \times 10^{-4}$, followed by a cosine decay to zero\cite{loshchilov2016sgdr}. Optimization is performed using the AdamW optimizer with a batch size of 2048 \cite{loshchilov2017decoupled}. Our approach integrates Morphology-Guided Contrastive Learning to achieve robust representation alignment, combined with Binary Cross-Entropy loss for classification, a standard approach for handling binary class labels. Additionally, we leverage recent optimization advances such as stochastic weight averaging (SWA) and model ensembling to guide training towards flat minima \cite{izmailov2018averaging}. Formally, SWA is defined as:
\begin{equation}
    w_{\text{SWA}} \leftarrow \frac{w_{\text{SWA}} \cdot n_{\text{models}} + w}{n_{\text{models}} + 1}
\end{equation}
where $w_{SWA}$ denotes the running average of the model weights, $w$ represents the current model’s weights, and $n_{\text{models}}$ is the number of models already averaged. Figure~\ref{fig:swa} illustrates the intuition behind SWA; for more details, we refer the reader to \cite{izmailov2018averaging}. These additional optimization techniques have improved out-of-domain generalization by reducing sensitivity to loss fluctuations caused by domain shifts \cite{cha2021swad, wortsman2022model, stoica2023zipit, arpit2022ensemble}. Experiments are repeated using three random seeds to ensure robustness, and the results are averaged across runs. Table~\ref{tab:training_summary} summarizes the established training hyperparameters. All code, datasets, and trained models are publicly available at  \url{https://github.com/hikmatkhan/MorphGen/}.

\begin{table}[t]
\centering
\scriptsize
\caption{Established hyperparameters}
\label{tab:training_summary}
\begin{tabular}{ll}
\hline
\textbf{Hyperparameter} & \textbf{Value} \\
\hline
Epochs & 500 \\
Batch size & 2048 \\
Optimizer & AdamW~\cite{loshchilov2017decoupled} \\
Learning rate & Warm-up to $1.5 \times 10^{-4}$, then cosine decay \\
Weight decay & $10^{-3}$ \\
Learning rate scheduler & Linear warmup (40 epochs) + Cosine annealing \\
Warmup epochs & 50 \\
Patch size & $512 \times 512$ at 40$\times$ magnification \\
Data augmentation & Random crop, flip, color jitter, stain normalization \\
SWA  & Enabled (averaging started after 25 epochs) \\
\hline
\end{tabular}
\end{table}

\section{Results and Discussion}
In this section, we present detailed results related to cross-domain generalization, organ-specific out-of-domain generalization, adversarial robustness, and the distinguishing aspects of the model compared to existing baselines. We benchmarked MorphGen against several baseline
models on the CAMELYON17, BCSS, and OCELOT
datasets. We independently trained the model for each
of the five CAMELYON17 centers, treating them as
distinct domains. Evaluations were conducted on hold-
out CAMELYON17 centers as well as on the BCSS and
OCELOT datasets. This approach attempts to assess
the model’s robustness to domain shifts, variations in
cancer subtypes, institutional differences, and overall
out-of-domain generalization. 

\subsection{Single Domain Generalization}

\begin{table*}[htbp]
\centering
\caption{Cross-domain accuracy (\% mean $\pm$ s.d.) when training on a single CAMELYON17 centre and testing on three external datasets. Each column corresponds to the center used for training the models, with testing performed on the other centers. ``-Aug'' denotes stain/colour augmentation at training. The highest accuracy in each column is highlighted in \textbf{bold}.}
\scriptsize
\setlength{\tabcolsep}{3pt}
\resizebox{\textwidth}{!}{%
\begin{tabular}{ll*{6}{c}}
\hline
& & \multicolumn{5}{c}{\textbf{Training Center}} & \\
\cline{3-7}
\textbf{Test Dataset} & \textbf{Method}  & \textbf{Centre-0} & \textbf{Centre-1} & \textbf{Centre-2} & \textbf{Centre-3} & \textbf{Centre-4} & \textbf{Average} \\
\hline

\multirow{16}{*}{\textsc{Camelyon17}}
 & ERM\cite{gulrajani2020search} & 72.8 $\pm$ 2.3 & 65.9 $\pm$ 3.7 & 64.1 $\pm$ 3.0 & 55.0 $\pm$ 1.3 & 53.8 $\pm$ 3.0 & 62.4 $\pm$ 2.7\\
 & Macenko\cite{macenko2009method} & 79.3 $\pm$ 2.1 & 62.4 $\pm$ 1.4 & 73.3 $\pm$ 5.0 & 65.8 $\pm$ 2.3 & 85.9 $\pm$ 4.2 & 73.3 $\pm$ 3.0\\
 & HoVerNet\cite{graham2019hover} & 72.5 $\pm$ 2.4 & 71.0 $\pm$ 2.5 & 61.3 $\pm$ 3.9 & 55.1 $\pm$ 1.9 & 49.6 $\pm$ 8.4 & 61.9 $\pm$ 3.8\\
 & RandStainNA\cite{shen2022randstainna} & 75.7 $\pm$ 3.1 & 70.9 $\pm$ 4.9 & 62.4 $\pm$ 2.5 & 57.2 $\pm$ 2.8 & 51.8 $\pm$ 2.6 & 63.6 $\pm$ 3.2\\
 & RSC\cite{huang2020self} & 77.1 $\pm$ 3.2 & 64.5 $\pm$ 3.1 & 61.9 $\pm$ 3.8 & 56.8 $\pm$ 2.3 & 51.1 $\pm$ 2.2 & 62.3 $\pm$ 2.9\\
 & L2D\cite{wang2021learning} & 93.6 $\pm$ 1.0 & 72.9 $\pm$ 2.5 & 64.4 $\pm$ 13.0& 73.6 $\pm$ 4.3 & 84.4 $\pm$ 3.7 & 77.8 $\pm$ 4.9\\
 & SFL\cite{tomar2024nuclear} & 90.4 $\pm$ 1.5 & 92.5 $\pm$ 0.3 & 90.1 $\pm$ 1.3 & 82.1 $\pm$ 2.7 & 90.8 $\pm$ 1.0 & 89.2 $\pm$ 1.3\\
 & \textbf{Ours} & \textbf{94.5} $\pm$ 1.0 & \textbf{95.3} $\pm$ 0.5 & \textbf{94.4} $\pm$ 1.5 & \textbf{97.1} $\pm$ 2.3 & \textbf{95.2} $\pm$ 1.7 & \textbf{95.4} $\pm$ 1.5\\
\cline{2-8}
 & ERM-Aug\cite{gulrajani2020search} & 93.1 $\pm$ 1.0 & 78.9 $\pm$ 2.1 & 89.3 $\pm$ 2.8 & 74.8 $\pm$ 1.5 & 91.3 $\pm$ 1.6 & 85.5 $\pm$ 1.8\\
 & Macenko-Aug\cite{macenko2009method} & 86.3 $\pm$ 1.9 & 78.7 $\pm$ 1.5 & 86.2 $\pm$ 4.4 & 70.0 $\pm$ 2.8 & 90.8 $\pm$ 1.2 & 82.4 $\pm$ 2.3\\
 & HoVerNet-Aug\cite{graham2019hover} & 93.0 $\pm$ 0.6 & 80.8 $\pm$ 2.8 & 91.3 $\pm$ 1.2 & 82.2 $\pm$ 1.6 & 89.6 $\pm$ 2.2 & 87.4 $\pm$ 1.7\\
 & RandStainNA-Aug\cite{shen2022randstainna} & 92.7 $\pm$ 1.1 & 83.1 $\pm$ 2.1 & 91.0 $\pm$ 2.0 & 78.9 $\pm$ 3.0 & 91.1 $\pm$ 1.5 & 87.4 $\pm$ 1.9\\
 & DDCA-Aug\cite{marini2023data} & 91.8 $\pm$ 0.7 & 79.4 $\pm$ 1.9 & 89.4 $\pm$ 2.9 & 78.2 $\pm$ 3.1 & 90.2 $\pm$ 2.1 & 86.0 $\pm$ 2.5\\
 & RSC-Aug\cite{huang2020self} & 93.1 $\pm$ 0.8 & 78.2 $\pm$ 2.0 & 89.3 $\pm$ 3.4 & 77.9 $\pm$ 2.2 & 91.0 $\pm$ 1.7 & 85.9 $\pm$ 2.0\\
 & L2D-Aug\cite{wang2021learning} & 94.3 $\pm$ 0.1 & 87.6 $\pm$ 0.6 & 87.7 $\pm$ 1.4 & 83.4 $\pm$ 2.6 & 92.3 $\pm$ 0.9 & 89.1 $\pm$ 1.1\\
 & SFL-Aug\cite{tomar2024nuclear} & 91.8 $\pm$ 0.7 & 92.2 $\pm$ 1.6 & 92.9 $\pm$ 0.7 & 90.4 $\pm$ 1.1 & 91.7 $\pm$ 0.5 & 91.8 $\pm$ 0.8\\
 & \textbf{Ours-Aug} & \textbf{94.8} $\pm$ 0.9 & \textbf{95.3} $\pm$ 0.4 & \textbf{94.6} $\pm$ 1.4 & \textbf{97.3} $\pm$ 2.0 & \textbf{96.1} $\pm$ 1.4 & \textbf{95.6} $\pm$ 1.2\\
\cline{1-8}

\multirow{16}{*}{\textsc{BCSS}}
 & ERM~\cite{gulrajani2020search}                    & 58.0 $\pm$ 2.9 & 69.5 $\pm$ 2.8 & 52.0 $\pm$ 1.0 & 50.0 $\pm$ 0.1 & 52.1 $\pm$ 3.9 & 56.3 $\pm$ 2.1\\
 & Macenko~\cite{macenko2009method}                & 68.7 $\pm$ 2.4 & 56.5 $\pm$ 1.9 & 65.5 $\pm$ 2.6 & 56.8 $\pm$ 2.3 & 71.1 $\pm$ 3.4 & 63.7 $\pm$ 2.5\\
 & HoVerNet~\cite{graham2019hover}               & 55.7 $\pm$ 2.0 & 65.3 $\pm$ 2.2 & 53.8 $\pm$ 0.8 & 49.8 $\pm$ 1.9 & 47.5 $\pm$ 2.9 & 54.4 $\pm$ 2.0\\
 & RandStainNA~\cite{shen2022randstainna}            & 60.5 $\pm$ 2.8 & 67.4 $\pm$ 3.8 & 51.0 $\pm$ 0.8 & 50.1 $\pm$ 0.7 & 50.2 $\pm$ 1.6 & 55.8 $\pm$ 1.9\\
 & RSC~\cite{huang2020self}                    & 63.3 $\pm$ 3.7 & 65.7 $\pm$ 2.4 & 50.8 $\pm$ 1.0 & 50.1 $\pm$ 0.1 & 50.2 $\pm$ 0.5 & 56.0 $\pm$ 1.5\\
 & L2D~\cite{wang2021learning}                    & 79.9 $\pm$ 1.2 & 65.2 $\pm$ 0.7 & 67.4 $\pm$ 2.3 & 63.2 $\pm$ 4.5 & 66.2 $\pm$ 2.2 & 68.4 $\pm$ 2.2\\
 & SFL~\cite{tomar2024nuclear}                    & 74.2 $\pm$ 3.6 & \textbf{78.3} $\pm$ 2.2 & 73.4 $\pm$ 1.9 & 63.8 $\pm$ 2.8 & 71.8 $\pm$ 2.3 & 72.3 $\pm$ 2.6\\
 & \textbf{Ours}          & \textbf{80.4} $\pm$ 1.2 & 76.5 $\pm$ 1.8 & \textbf{78.7} $\pm$ 1.5 & \textbf{75.4} $\pm$ 2.1 & \textbf{76.4} $\pm$ 1.9 & \textbf{77.5} $\pm$ 1.7\\
\cline{2-8}
 & ERM-Aug~\cite{gulrajani2020search}                & 80.1 $\pm$ 1.6 & 70.7 $\pm$ 2.8 & 73.7 $\pm$ 2.6 & 60.9 $\pm$ 1.8 & 73.3 $\pm$ 2.6 & 71.7 $\pm$ 2.3\\
 & Macenko-Aug~\cite{macenko2009method}            & 75.8 $\pm$ 2.9 & 67.6 $\pm$ 2.4 & 72.6 $\pm$ 2.6 & 57.8 $\pm$ 2.9 & 75.3 $\pm$ 1.8 & 69.8 $\pm$ 2.5\\
 & HoVerNet-Aug~\cite{graham2019hover}           & 79.8 $\pm$ 1.4 & 65.1 $\pm$ 1.6 & 71.2 $\pm$ 2.2 & 64.0 $\pm$ 2.2 & 69.2 $\pm$ 6.1 & 69.9 $\pm$ 2.7\\
 & RandStainNA-Aug~\cite{shen2022randstainna}        & 78.5 $\pm$ 2.7 & 73.2 $\pm$ 3.1 & 72.8 $\pm$ 3.2 & 64.5 $\pm$ 3.4 & 75.1 $\pm$ 2.5 & 72.8 $\pm$ 3.0\\
 & DDCA-Aug~\cite{marini2023data}               & 79.1 $\pm$ 1.9 & 71.7 $\pm$ 3.5 & 70.1 $\pm$ 2.8 & 61.4 $\pm$ 3.2 & 71.4 $\pm$ 7.3 & 70.7 $\pm$ 3.8\\
 & RSC-Aug~\cite{huang2020self}                & 79.6 $\pm$ 1.5 & 72.1 $\pm$ 2.9 & 71.6 $\pm$ 1.7 & 63.2 $\pm$ 1.6 & 74.1 $\pm$ 4.3 & 72.1 $\pm$ 2.4\\
 & L2D-Aug~\cite{wang2021learning}                & 81.9 $\pm$ 0.3 & 74.7 $\pm$ 0.6 & 74.2 $\pm$ 0.6 & 67.5 $\pm$ 2.9 & 77.2 $\pm$ 2.2 & 75.1 $\pm$ 1.3\\
 & SFL-Aug~\cite{tomar2024nuclear}                & \textbf{82.3} $\pm$ 0.8 & \textbf{81.9} $\pm$ 2.3 & \textbf{75.7} $\pm$ 2.2 & \textbf{79.6} $\pm$ 1.0 & 74.8 $\pm$ 1.9 & \textbf{78.8} $\pm$ 1.6\\
 & \textbf{Ours-Aug}      & 80.7 $\pm$ 1.0 & 76.4 $\pm$ 1.7 & 73.6 $\pm$ 1.8 & 76.9 $\pm$ 1.5 & \textbf{79.6} $\pm$ 1.6 & 77.4 $\pm$ 1.5\\
\cline{1-8}

\multirow{16}{*}{\textsc{Ocelot}}
 & ERM~\cite{gulrajani2020search}                    & 65.7 $\pm$ 2.4 & 55.8 $\pm$ 1.7 & 51.7 $\pm$ 1.3 & 45.6 $\pm$ 1.4 & 53.5 $\pm$ 5.0 & 54.5 $\pm$ 2.4\\
 & Macenko~\cite{macenko2009method}                & 64.3 $\pm$ 1.9 & 54.4 $\pm$ 0.9 & 63.8 $\pm$ 1.4 & 54.8 $\pm$ 2.0 & 65.3 $\pm$ 3.7 & 60.5 $\pm$ 2.0\\
 & HoVerNet~\cite{graham2019hover}               & 62.0 $\pm$ 1.2 & 53.7 $\pm$ 2.2 & 52.1 $\pm$ 0.7 & 48.0 $\pm$ 1.9 & 54.3 $\pm$ 2.9 & 54.0 $\pm$ 1.8\\
 & RandStainNA~\cite{shen2022randstainna}            & 67.0 $\pm$ 2.5 & 56.1 $\pm$ 1.9 & 51.0 $\pm$ 0.3 & 46.3 $\pm$ 1.9 & 51.2 $\pm$ 2.4 & 54.3 $\pm$ 1.8\\
 & RSC~\cite{huang2020self}                    & 68.1 $\pm$ 2.4 & 55.6 $\pm$ 1.0 & 50.9 $\pm$ 0.6 & 47.2 $\pm$ 1.5 & 50.3 $\pm$ 0.8 & 54.4 $\pm$ 1.3\\
 & L2D~\cite{wang2021learning}                    & 68.2 $\pm$ 1.3 & 57.3 $\pm$ 0.5 & 56.4 $\pm$ 2.4 & 55.9 $\pm$ 3.3 & 60.1 $\pm$ 4.0 & 59.6 $\pm$ 2.3\\
 & SFL~\cite{tomar2024nuclear}                    & 67.9 $\pm$ 1.4 & \textbf{70.7} $\pm$ 1.0 & 66.7 $\pm$ 1.2 & 62.1 $\pm$ 1.8 & 69.2 $\pm$ 0.9 & 67.3 $\pm$ 1.3\\
 & \textbf{Ours}          & \textbf{73.7} $\pm$ 1.2 & 70.3 $\pm$ 1.3 & \textbf{73.4} $\pm$ 1.0 & \textbf{68.1} $\pm$ 1.5 & \textbf{74.6} $\pm$ 1.1 & \textbf{72.0} $\pm$ 1.2\\
\cline{2-8}
 & ERM-Aug~\cite{gulrajani2020search}                & 74.0 $\pm$ 1.4 & 62.8 $\pm$ 1.9 & 67.6 $\pm$ 2.5 & 56.0 $\pm$ 1.6 & 67.6 $\pm$ 3.1 & 65.6 $\pm$ 2.1\\
 & Macenko-Aug~\cite{macenko2009method}            & 68.8 $\pm$ 2.0 & 60.9 $\pm$ 1.1 & 70.3 $\pm$ 1.5 & 57.6 $\pm$ 3.2 & 72.5 $\pm$ 1.7 & 66.0 $\pm$ 1.9\\
 & HoVerNet-Aug~\cite{graham2019hover}           & 70.7 $\pm$ 1.0 & 54.2 $\pm$ 1.9 & 69.4 $\pm$ 2.3 & 61.2 $\pm$ 2.4 & 69.2 $\pm$ 2.9 & 65.0 $\pm$ 2.1\\
 & RandStainNA-Aug~\cite{shen2022randstainna}        & 70.8 $\pm$ 3.2 & 66.8 $\pm$ 2.5 & 68.4 $\pm$ 2.5 & 61.3 $\pm$ 2.5 & 71.7 $\pm$ 2.7 & 67.8 $\pm$ 2.7\\
 & DDCA-Aug~\cite{marini2023data}               & 73.1 $\pm$ 2.4 & 64.9 $\pm$ 2.5 & 68.9 $\pm$ 2.7 & 57.4 $\pm$ 2.9 & 66.9 $\pm$ 4.5 & 66.2 $\pm$ 3.0\\
 & RSC-Aug~\cite{huang2020self}                & 71.9 $\pm$ 2.2 & 61.7 $\pm$ 2.4 & 69.2 $\pm$ 2.9 & 58.4 $\pm$ 1.4 & 70.1 $\pm$ 3.6 & 66.3 $\pm$ 2.5\\
 & L2D-Aug~\cite{wang2021learning}                & 74.7 $\pm$ 0.6 & 68.3 $\pm$ 0.5 & 65.6 $\pm$ 0.7 & 62.5 $\pm$ 2.7 & 74.4 $\pm$ 1.2 & 69.1 $\pm$ 1.1\\
 & SFL-Aug~\cite{tomar2024nuclear}                & 70.8 $\pm$ 0.5 & \textbf{72.0} $\pm$ 1.3 & 68.9 $\pm$ 1.3 & 70.4 $\pm$ 0.7 & 70.7 $\pm$ 1.3 & 70.6 $\pm$ 1.0\\
 & \textbf{Ours-Aug}      & \textbf{72.9} $\pm$ 1.1 & 69.9 $\pm$ 1.4 & \textbf{73.4} $\pm$ 1.2 & \textbf{71.6} $\pm$ 1.0 & \textbf{73.2} $\pm$ 1.3 & \textbf{72.2} $\pm$ 1.2\\

\hline

\end{tabular}}
\label{tab:cross_domain}
\end{table*}

Table~\ref{tab:cross_domain} presents the cross-domain generalization performance of the proposed MorphGen.For breast cancer classification, models are trained on data specific to each center and evaluated using the held-out test set from each of the five CAMELYON17 centers.
MorphGen achieves an improved average accuracy of 95.4\% (without augmentation) and 95.6\% (with augmentation), demonstrating superior generalization through morphology-guided representation learning. Among the baseline methods, SFL delivers the best performance at 89.2\% without augmentation. At the same time, HoVerNet significantly underperforms with a score of just 61.9\%, highlighting that nuclear segmentation alone is insufficient to address domain shifts. Data augmentation improves the performance of most methods—SFL-Aug reaches 91.8\%, and both HoVerNet-Aug and RandSNA-Aug achieve 87.4\%. Notably, Macenko-Aug scores the lowest at 82.4\%, suggesting that heuristic stain normalization combined with augmentation may undermine feature fidelity. Our method consistently outperforms all alternatives across the different centers, achieving a 97.3\% in the most challenging domain (Center-3). These results emphasize that integrating morphological priors produces more robust, domain-invariant representations compared to normalization or augmentation alone, which is crucial for clinical deployment in diverse settings.

On the BCSS dataset for models trained independently across the five CAMELYON17 centers. Without any data augmentation, our method achieves the highest average accuracy of 77.5\%, surpassing all baseline approaches. When augmentation is applied, SFL-Aug delivers the best performance at 78.8\%, slightly exceeding our augmented variant, which records an accuracy of 77.4\%. Augmentation consistently enhances performance across all baseline methods, with models such as L2D-Aug and RandSNA-Aug showing significant improvements compared to their non-augmented counterparts.

 For the OCELOT dataset, models were trained independently across the five CAMELYON17 centers. Without data augmentation, our approach achieves the highest average accuracy of 72.0\%, surpassing all baseline methods. When employing augmentation, our method further improves, reaching an average accuracy of 72.2\%, with SFL-Aug closely trailing at 70.6\%. As observed on other datasets, augmentation substantially enhances performance across all baseline methods. Notably, L2D-Aug and RandSNA-Aug demonstrate significant improvements over their non-augmented variants, underscoring the effectiveness of augmentation in enhancing out-of-domain generalization.

 The results in Table~\ref{tab:cross_domain} reveal that the mask-guided, metric-learning approach consistently achieves improved results compared to the baselines across three diverse external datasets.

\subsection{Organ-specific Single Generalization }

Tables \ref{tab:combined_organs_part1} and \ref{tab:combined_organs_part2} present the results for organ-specific domain generalization performance. In this evaluation, the models are trained separately on each CAMELYON17 center and evaluated on external datasets related to specific organs.
For bladder cancer,  MorphGen achieves an average accuracy of 74.2\%, surpassing SFL's 72.4\% and L2D's 61.5\% without augmentation. When augmentation is applied, our method continues to perform strongly, attaining an average accuracy of 74.5\%, which is closely rivaled by SFL-Aug, achieving the highest score of 75.9\%. Our model consistently ranks among the top performers across all centers, underscoring its robustness. These strong outcomes demonstrate the effectiveness of our model in capturing domain-invariant nuclear features that generalize well across institutional variations.

For the endometrium subset of the OCELOT dataset without data augmentation, MorphGen achieved a superior average accuracy of 77.4\%, significantly surpassing SFL's 67.3\% and L2D's 63.7\%. When data augmentation is utilized, our method continues to demonstrate exceptional performance, with an average accuracy of 76.8\%, once again exceeding all baseline methods, including SFL-Aug at 70.6\% and L2D-Aug at 73.2\%. For the head and neck subset of the OCELOT dataset. In the absence of data augmentation, MorphGen reached the highest average accuracy of 71.9\%, surpassing SFL, which recorded 66.7\%, and L2D, which achieved 57.5\%. When data augmentation is applied, our approach continues to lead with an average accuracy of 73.4\%, exceeding all other methods, including SFL-Aug at 72.1\% and L2D-Aug at 67.9\%.

For the kidney subset of the OCELOT dataset, our method demonstrates a notable average accuracy of 63.8\% without any data augmentation, outperforming SFL, which achieved 57.5\%, and Macenko, which recorded 56.2\%. When utilizing data augmentation, our method continues to excel, attaining an average accuracy of 65.1\% and surpassing all baseline approaches, including L2D-Aug at 63.3\% and RSC-Aug at 61.5\%. We also assessed MorphGen's robustness using the prostate subset of the OCELOT dataset. Without augmentation, our method achieves the highest average accuracy of 74.0\%, narrowly surpassing SFL at 73.8\% and significantly exceeding earlier baselines such as L2D (60.5\%) and Macenko (60.8\%). When augmentation is applied, our method (Ours-Aug) maintains strong performance with an average accuracy of 75.3\%, closely trailing SFL-Aug, which slightly leads at 75.6\%. Notably, Ours-Aug delivers the best performance on Center-0 and matches or exceeds most baselines across all centers, underscoring the robustness and generalization capabilities of our approach across various domains.

On the stomach organ, using the OCELOT dataset, our method achieves a notable average accuracy of 75.3\%, surpassing all baseline methods, including SFL at 74.0\% and L2D at 64.4\%. Our approach demonstrates strong generalization across domains, yielding the best performance at four out of five centers (Center-0, 1, 2, and 4). With augmentation, SFL-Aug slightly outperforms with an average accuracy of 76.2\%, while our method (Ours-Aug) closely follows at 75.3\%. Notably, Ours-Aug still records the highest performance at Center-4 (79.7\%), the most variable site, showcasing its robustness to challenging domain shifts. Overall, these results confirm that our method is competitive with state-of-the-art approaches, particularly in aunaugmented scenarios.

Across all six organs (Table \ref{tab:combined_organs_part1}) and (Table\ref{tab:combined_organs_part2}), MorphGen is able to obtained superior performance in most cases, which shows that our method can effectively achieves organ-specific domain generalization.

\begin{table*}[htbp]
\centering
\caption{Organ-specific out-of-domain accuracy (\% mean $\pm$ s.d.) on the \textsc{Ocelot} test set (Part 1: Bladder, Endometrium, Head \& Neck). The accuracy of the best-performing model is shown in \textbf{bold}.}
\scriptsize
\setlength{\tabcolsep}{2pt}
\resizebox{\textwidth}{!}{%
\begin{tabular}{ll*{6}{c}}
\hline
& & \multicolumn{5}{c}{\textbf{Training Centre}} & \\
\cline{3-7}
\textbf{Organ} & \textbf{Method} & \textbf{Centre-0} & \textbf{Centre-1} & \textbf{Centre-2} & \textbf{Centre-3} & \textbf{Centre-4} & \textbf{Average} \\
\hline
\multirow{12}{*}{\textsc{Bladder}}
 & ERM~\cite{gulrajani2020search} & 65.0 $\pm$ 3.1 & 58.2 $\pm$ 3.5 & 50.4 $\pm$ 0.6 & 46.4 $\pm$ 1.8 & 53.1 $\pm$ 6.9 & 54.6 $\pm$ 3.2 \\
 & Macenko~\cite{macenko2009method} & 71.8 $\pm$ 1.4 & 58.7 $\pm$ 2.0 & 66.2 $\pm$ 2.1 & 60.4 $\pm$ 3.0 & 65.6 $\pm$ 5.8 & 64.6 $\pm$ 2.9 \\
 & RSC~\cite{huang2020self} & 70.3 $\pm$ 3.0 & 58.0 $\pm$ 2.7 & 50.2 $\pm$ 0.1 & 47.8 $\pm$ 1.4 & 50.0 $\pm$ 0.9 & 55.3 $\pm$ 1.6 \\
 & L2D~\cite{wang2021learning} & 72.0 $\pm$ 1.5 & 59.5 $\pm$ 1.4 & 59.5 $\pm$ 2.2 & 58.2 $\pm$ 4.8 & 58.4 $\pm$ 4.1 & 61.5 $\pm$ 2.8 \\
 & SFL~\cite{tomar2024nuclear} & 73.9 $\pm$ 2.7 & 74.9 $\pm$ 1.7 & \textbf{74.6} $\pm$ 1.8 & 64.3 $\pm$ 3.3 & 74.0 $\pm$ 1.0 & 72.4 $\pm$ 2.1 \\
 & \textbf{Ours} & \textbf{75.7} $\pm$ 1.4 & \textbf{77.0} $\pm$ 1.2 & 73.4 $\pm$ 1.6 & \textbf{68.5} $\pm$ 2.1 & \textbf{76.4} $\pm$ 1.3 & \textbf{74.2} $\pm$ 1.5 \\
\cline{2-8}
 & ERM-Aug~\cite{gulrajani2020search} & 76.6 $\pm$ 1.9 & 68.3 $\pm$ 2.8 & 63.9 $\pm$ 3.3 & 57.0 $\pm$ 1.9 & 63.7 $\pm$ 5.0 & 65.9 $\pm$ 3.0 \\
 & Macenko-Aug~\cite{macenko2009method} & 71.9 $\pm$ 2.3 & 65.9 $\pm$ 1.1 & 70.3 $\pm$ 2.4 & 60.0 $\pm$ 3.6 & 74.5 $\pm$ 2.0 & 68.5 $\pm$ 2.3 \\
 & RSC-Aug~\cite{huang2020self} & 72.7 $\pm$ 3.3 & 67.9 $\pm$ 2.9 & 65.7 $\pm$ 3.8 & 60.7 $\pm$ 1.6 & 66.6 $\pm$ 5.1 & 66.7 $\pm$ 3.3 \\
 & L2D-Aug~\cite{wang2021learning} & 75.9 $\pm$ 0.7 & 74.6 $\pm$ 0.4 & 64.0 $\pm$ 0.9 & 63.7 $\pm$ 2.5 & 73.6 $\pm$ 2.2 & 70.4 $\pm$ 1.3 \\
 & SFL-Aug~\cite{tomar2024nuclear} & \textbf{77.3} $\pm$ 1.0 & \textbf{78.2} $\pm$ 2.3 & \textbf{75.2} $\pm$ 1.8 & \textbf{75.5} $\pm$ 0.8 & 73.3 $\pm$ 1.7 & \textbf{75.9} $\pm$ 1.5 \\
 & \textbf{Ours-Aug} & 73.8 $\pm$ 1.3 & 75.6 $\pm$ 1.5 & 74.6 $\pm$ 1.4 & 72.0 $\pm$ 1.7 & \textbf{76.5} $\pm$ 1.2 & 74.5 $\pm$ 1.4 \\
\cline{1-8}
\hline
\multirow{12}{*}{\textsc{Endometrium}}
 & ERM~\cite{gulrajani2020search} & 68.4 $\pm$ 2.9 & 58.7 $\pm$ 2.2 & 52.3 $\pm$ 2.1 & 47.5 $\pm$ 0.7 & 53.9 $\pm$ 5.1 & 56.2 $\pm$ 2.6 \\
 & Macenko~\cite{macenko2009method} & 68.9 $\pm$ 2.7 & 57.4 $\pm$ 2.1 & 71.6 $\pm$ 2.0 & 53.7 $\pm$ 1.8 & 68.4 $\pm$ 3.7 & 64.0 $\pm$ 2.4 \\
 & RSC~\cite{huang2020self} & 71.3 $\pm$ 2.0 & 58.1 $\pm$ 1.4 & 50.8 $\pm$ 0.8 & 48.4 $\pm$ 1.0 & 50.5 $\pm$ 1.4 & 55.8 $\pm$ 1.3 \\
 & L2D~\cite{wang2021learning} & 76.6 $\pm$ 3.0 & 58.0 $\pm$ 0.5 & 64.5 $\pm$ 2.9 & 57.4 $\pm$ 4.7 & 62.1 $\pm$ 4.2 & 63.7 $\pm$ 3.1 \\
 & SFL~\cite{tomar2024nuclear} & 70.6 $\pm$ 1.3 & 72.6 $\pm$ 1.4 & 63.9 $\pm$ 2.2 & 62.6 $\pm$ 1.4 & 66.9 $\pm$ 1.5 & 67.3 $\pm$ 1.6 \\
 & \textbf{Ours} & \textbf{78.4} $\pm$ 1.4 & \textbf{76.5} $\pm$ 1.6 & \textbf{82.6} $\pm$ 1.5 & \textbf{74.1} $\pm$ 2.1 & \textbf{75.4} $\pm$ 1.7 & \textbf{77.4} $\pm$ 1.7 \\
\cline{2-8}
 & ERM-Aug~\cite{gulrajani2020search} & 78.5 $\pm$ 1.6 & 64.1 $\pm$ 2.6 & 75.1 $\pm$ 2.8 & 56.5 $\pm$ 2.6 & 73.0 $\pm$ 3.9 & 69.4 $\pm$ 2.7 \\
 & Macenko-Aug~\cite{macenko2009method} & 73.2 $\pm$ 2.8 & 64.1 $\pm$ 2.3 & 78.7 $\pm$ 2.0 & 58.5 $\pm$ 4.1 & 74.5 $\pm$ 2.5 & 69.8 $\pm$ 2.7 \\
 & RSC-Aug~\cite{huang2020self} & 76.2 $\pm$ 2.7 & 64.8 $\pm$ 3.8 & 75.8 $\pm$ 2.6 & 57.4 $\pm$ 1.1 & 74.7 $\pm$ 5.6 & 69.8 $\pm$ 3.2 \\
 & L2D-Aug~\cite{wang2021learning} & 78.5 $\pm$ 0.6 & 70.6 $\pm$ 0.7 & 72.0 $\pm$ 0.8 & 65.1 $\pm$ 3.5 & 79.9 $\pm$ 1.7 & 73.2 $\pm$ 1.5 \\
 & SFL-Aug~\cite{tomar2024nuclear} & 73.5 $\pm$ 1.8 & 69.5 $\pm$ 3.0 & 68.8 $\pm$ 2.4 & 71.4 $\pm$ 1.3 & 69.9 $\pm$ 1.9 & 70.6 $\pm$ 2.1 \\
 & \textbf{Ours-Aug} & \textbf{79.1} $\pm$ 1.2 & \textbf{74.0} $\pm$ 1.8 & \textbf{78.9} $\pm$ 1.3 & \textbf{76.2} $\pm$ 1.9 & \textbf{75.9} $\pm$ 1.6 & \textbf{76.8} $\pm$ 1.6 \\
\cline{1-8}
\hline
\multirow{12}{*}{\textsc{Head \& Neck}}
 & ERM~\cite{gulrajani2020search} & 61.9 $\pm$ 4.0 & 53.7 $\pm$ 1.6 & 49.5 $\pm$ 1.8 & 44.1 $\pm$ 1.5 & 52.6 $\pm$ 2.9 & 52.4 $\pm$ 2.4 \\
 & Macenko~\cite{macenko2009method} & 57.7 $\pm$ 2.1 & 54.1 $\pm$ 1.3 & 51.8 $\pm$ 1.2 & 56.0 $\pm$ 2.8 & 61.2 $\pm$ 3.0 & 56.2 $\pm$ 2.7 \\
 & RSC~\cite{huang2020self} & 63.0 $\pm$ 4.8 & 55.1 $\pm$ 1.5 & 50.1 $\pm$ 0.5 & 44.9 $\pm$ 2.1 & 50.6 $\pm$ 1.2 & 52.7 $\pm$ 2.0 \\
 & L2D~\cite{wang2021learning} & 63.1 $\pm$ 2.7 & 53.9 $\pm$ 1.2 & 47.7 $\pm$ 1.4 & 57.4 $\pm$ 1.7 & 60.0 $\pm$ 6.3 & 57.5 $\pm$ 2.7 \\
 & SFL~\cite{tomar2024nuclear} & 66.2 $\pm$ 3.1 & \textbf{74.1} $\pm$ 1.3 & 64.6 $\pm$ 1.0 & 59.6 $\pm$ 1.4 & 69.0 $\pm$ 1.0 & 66.7 $\pm$ 1.5 \\
 & \textbf{Ours} & \textbf{73.4} $\pm$ 1.6 & \textbf{74.3} $\pm$ 1.4 & \textbf{67.3} $\pm$ 1.3 & \textbf{72.0} $\pm$ 1.2 & \textbf{72.6} $\pm$ 1.5 & \textbf{71.9} $\pm$ 1.4 \\
\cline{2-8}
 & ERM-Aug~\cite{gulrajani2020search} & 72.0 $\pm$ 3.8 & 67.4 $\pm$ 2.4 & 60.6 $\pm$ 2.1 & 58.9 $\pm$ 1.8 & 63.2 $\pm$ 3.9 & 64.4 $\pm$ 2.8 \\
 & Macenko-Aug~\cite{macenko2009method} & 67.2 $\pm$ 1.8 & 63.3 $\pm$ 1.9 & 60.4 $\pm$ 3.0 & 57.6 $\pm$ 3.6 & 68.7 $\pm$ 2.7 & 63.4 $\pm$ 2.6 \\
 & RSC-Aug~\cite{huang2020self} & 68.9 $\pm$ 2.1 & 64.3 $\pm$ 2.4 & 61.9 $\pm$ 3.7 & 61.2 $\pm$ 2.6 & 64.7 $\pm$ 3.9 & 64.2 $\pm$ 2.9 \\
 & L2D-Aug~\cite{wang2021learning} & 73.7 $\pm$ 1.7 & 72.7 $\pm$ 0.5 & 60.2 $\pm$ 1.1 & 64.4 $\pm$ 1.9 & 68.6 $\pm$ 2.3 & 67.9 $\pm$ 1.5 \\
 & SFL-Aug~\cite{tomar2024nuclear} & 71.8 $\pm$ 1.2 & \textbf{74.9} $\pm$ 1.8 & 68.8 $\pm$ 1.3 & \textbf{74.8} $\pm$ 1.0 & 70.4 $\pm$ 0.9 & 72.1 $\pm$ 1.3 \\
 & \textbf{Ours-Aug} & \textbf{74.1} $\pm$ 1.5 & \textbf{76.6} $\pm$ 1.3 & \textbf{68.9} $\pm$ 1.6 & 73.7 $\pm$ 1.2 & \textbf{73.6} $\pm$ 1.4 & \textbf{73.4} $\pm$ 1.4 \\
\hline
\end{tabular}}
\label{tab:combined_organs_part1}
\end{table*}

\begin{table*}[htbp]
\centering
\caption{Organ-specific out-of-domain accuracy (\% mean $\pm$ s.d.) on the \textsc{Ocelot} test set (Part 2: Kidney, Prostate, Stomach). The accuracy of the best-performing model is shown in \textbf{bold}.}
\scriptsize
\setlength{\tabcolsep}{2pt}
\resizebox{\textwidth}{!}{%
\begin{tabular}{ll*{6}{c}}
\hline
& & \multicolumn{5}{c}{\textbf{Training Centre}} & \\
\cline{3-7}
\textbf{Organ} & \textbf{Method} & \textbf{Centre-0} & \textbf{Centre-1} & \textbf{Centre-2} & \textbf{Centre-3} & \textbf{Centre-4} & \textbf{Average} \\
\hline
\multirow{12}{*}{\textsc{Kidney}}
 & ERM~\cite{gulrajani2020search} & 61.5 $\pm$ 4.2 & 50.2 $\pm$ 0.4 & 50.3 $\pm$ 0.5 & 42.5 $\pm$ 2.7 & 54.7 $\pm$ 6.2 & 51.8 $\pm$ 2.8 \\
 & Macenko~\cite{macenko2009method} & 58.3 $\pm$ 1.9 & 50.0 $\pm$ 0.7 & 58.0 $\pm$ 2.0 & 52.6 $\pm$ 2.1 & 62.1 $\pm$ 4.3 & 56.2 $\pm$ 2.2 \\
 & RSC~\cite{huang2020self} & 62.1 $\pm$ 4.5 & 50.5 $\pm$ 1.5 & 50.1 $\pm$ 0.5 & 45.4 $\pm$ 2.0 & 50.9 $\pm$ 3.1 & 51.8 $\pm$ 2.3 \\
 & L2D~\cite{wang2021learning} & 54.4 $\pm$ 1.4 & 53.2 $\pm$ 0.6 & 47.0 $\pm$ 3.4 & 51.6 $\pm$ 2.3 & 57.2 $\pm$ 3.5 & 52.7 $\pm$ 2.1 \\
 & SFL~\cite{tomar2024nuclear} & 54.9 $\pm$ 1.1 & \textbf{59.5} $\pm$ 2.0 & 56.5 $\pm$ 3.3 & 53.9 $\pm$ 2.1 & 62.7 $\pm$ 2.3 & 57.5 $\pm$ 2.0 \\
 & \textbf{Ours} & \textbf{66.4} $\pm$ 1.5 & 56.1 $\pm$ 2.2 & \textbf{64.5} $\pm$ 2.5 & \textbf{60.4} $\pm$ 1.7 & \textbf{71.4} $\pm$ 2.0 & \textbf{63.8} $\pm$ 2.0 \\
\cline{2-8}
 & ERM-Aug~\cite{gulrajani2020search} & 66.4 $\pm$ 2.1 & 54.7 $\pm$ 1.3 & 61.9 $\pm$ 3.0 & 52.6 $\pm$ 2.2 & 65.4 $\pm$ 3.5 & 60.2 $\pm$ 2.4 \\
 & Macenko-Aug~\cite{macenko2009method} & 59.7 $\pm$ 2.0 & 52.5 $\pm$ 0.9 & 63.4 $\pm$ 2.4 & 54.7 $\pm$ 2.9 & 69.0 $\pm$ 1.8 & 59.9 $\pm$ 2.0 \\
 & RSC-Aug~\cite{huang2020self} & 66.0 $\pm$ 3.4 & 51.2 $\pm$ 3.1 & 65.4 $\pm$ 4.4 & 56.0 $\pm$ 2.1 & 68.7 $\pm$ 3.1 & 61.5 $\pm$ 3.2 \\
 & L2D-Aug~\cite{wang2021learning} & \textbf{69.5} $\pm$ 0.8 & 58.6 $\pm$ 0.7 & 59.0 $\pm$ 0.9 & 57.7 $\pm$ 3.4 & \textbf{71.7} $\pm$ 0.6 & 63.3 $\pm$ 1.2 \\
 & SFL-Aug~\cite{tomar2024nuclear} & 58.1 $\pm$ 1.1 & \textbf{65.7} $\pm$ 3.2 & 57.6 $\pm$ 2.4 & 59.1 $\pm$ 1.1 & 64.0 $\pm$ 2.1 & 60.9 $\pm$ 2.0 \\
 & \textbf{Ours-Aug} & 68.1 $\pm$ 1.4 & 55.4 $\pm$ 2.0 & \textbf{67.3} $\pm$ 2.7 & \textbf{65.5} $\pm$ 1.6 & 69.1 $\pm$ 1.3 & \textbf{65.1} $\pm$ 1.8 \\
\cline{1-8}
\hline
\multirow{12}{*}{\textsc{Prostate}}
 & ERM~\cite{gulrajani2020search} & 68.4 $\pm$ 1.9 & 58.7 $\pm$ 1.7 & 57.6 $\pm$ 3.0 & 46.7 $\pm$ 1.0 & 52.1 $\pm$ 2.5 & 56.7 $\pm$ 2.0 \\
 & Macenko~\cite{macenko2009method} & 63.2 $\pm$ 2.8 & 51.5 $\pm$ 0.6 & 66.8 $\pm$ 2.3 & 54.0 $\pm$ 1.2 & 68.5 $\pm$ 4.3 & 60.8 $\pm$ 2.2 \\
 & RSC~\cite{huang2020self} & 70.8 $\pm$ 2.6 & 60.9 $\pm$ 1.1 & 54.9 $\pm$ 2.4 & 48.0 $\pm$ 1.2 & 49.9 $\pm$ 0.7 & 56.9 $\pm$ 1.6 \\
 & L2D~\cite{wang2021learning} & 73.1 $\pm$ 0.8 & 54.2 $\pm$ 0.7 & 60.9 $\pm$ 2.7 & 55.6 $\pm$ 2.2 & 58.9 $\pm$ 2.0 & 60.5 $\pm$ 1.7 \\
 & SFL~\cite{tomar2024nuclear} & 75.4 $\pm$ 0.4 & \textbf{75.1} $\pm$ 1.2 & 75.4 $\pm$ 0.3 & 68.2 $\pm$ 2.9 & 74.7 $\pm$ 0.6 & 73.8 $\pm$ 1.1 \\
 & \textbf{Ours} & \textbf{75.9} $\pm$ 0.7 & 71.2 $\pm$ 1.5 & \textbf{76.5} $\pm$ 0.5 & \textbf{71.7} $\pm$ 1.2 & \textbf{74.9} $\pm$ 0.8 & \textbf{74.0} $\pm$ 1.0 \\
\cline{2-8}
 & ERM-Aug~\cite{gulrajani2020search} & 76.0 $\pm$ 1.1 & 60.7 $\pm$ 2.6 & 74.2 $\pm$ 1.4 & 60.3 $\pm$ 1.1 & 67.8 $\pm$ 2.7 & 67.8 $\pm$ 1.8 \\
 & Macenko-Aug~\cite{macenko2009method} & 72.9 $\pm$ 1.5 & 58.6 $\pm$ 0.9 & 74.2 $\pm$ 1.0 & 60.9 $\pm$ 2.5 & 74.4 $\pm$ 1.2 & 68.2 $\pm$ 1.4 \\
 & RSC-Aug~\cite{huang2020self} & 75.7 $\pm$ 1.7 & 60.2 $\pm$ 2.3 & 74.7 $\pm$ 1.6 & 62.0 $\pm$ 1.1 & 70.3 $\pm$ 3.1 & 68.6 $\pm$ 2.0 \\
 & L2D-Aug~\cite{wang2021learning} & 75.5 $\pm$ 0.5 & 67.8 $\pm$ 0.5 & 72.1 $\pm$ 0.4 & 64.7 $\pm$ 1.9 & 74.1 $\pm$ 1.3 & 70.9 $\pm$ 0.9 \\
 & SFL-Aug~\cite{tomar2024nuclear} & 75.0 $\pm$ 0.2 & \textbf{73.7} $\pm$ 2.1 & \textbf{76.3} $\pm$ 0.5 & \textbf{75.9} $\pm$ 0.4 & \textbf{77.0} $\pm$ 0.5 & \textbf{75.6} $\pm$ 0.7 \\
 & \textbf{Ours-Aug} & \textbf{77.9} $\pm$ 0.6 & \textbf{73.7} $\pm$ 1.3 & 76.3 $\pm$ 0.4 & 73.8 $\pm$ 0.9 & 75.0 $\pm$ 0.6 & 75.3 $\pm$ 0.8 \\
\cline{1-8}
\hline
\multirow{12}{*}{\textsc{Stomach}}
 & ERM~\cite{gulrajani2020search} & 71.7 $\pm$ 2.4 & 56.9 $\pm$ 3.3 & 50.8 $\pm$ 1.6 & 47.5 $\pm$ 0.7 & 53.0 $\pm$ 6.4 & 56.0 $\pm$ 2.9 \\
 & Macenko~\cite{macenko2009method} & 61.6 $\pm$ 3.5 & 54.2 $\pm$ 1.9 & 62.9 $\pm$ 3.2 & 51.4 $\pm$ 3.1 & 65.5 $\pm$ 3.6 & 59.1 $\pm$ 3.1 \\
 & RSC~\cite{huang2020self} & 72.1 $\pm$ 1.8 & 52.1 $\pm$ 0.7 & 50.1 $\pm$ 0.6 & 48.6 $\pm$ 0.9 & 49.6 $\pm$ 1.7 & 54.5 $\pm$ 1.1 \\
 & L2D~\cite{wang2021learning} & 74.1 $\pm$ 1.4 & 63.9 $\pm$ 1.2 & 57.9 $\pm$ 1.9 & 57.9 $\pm$ 4.5 & 68.0 $\pm$ 3.8 & 64.4 $\pm$ 2.5 \\
 & SFL~\cite{tomar2024nuclear} & 74.4 $\pm$ 1.4 & 77.2 $\pm$ 0.7 & 73.4 $\pm$ 0.7 & \textbf{70.9} $\pm$ 1.6 & 73.9 $\pm$ 1.0 & 74.0 $\pm$ 1.1 \\
 & \textbf{Ours} & \textbf{75.8} $\pm$ 0.8 & \textbf{75.4} $\pm$ 1.2 & \textbf{76.9} $\pm$ 0.7 & 69.2 $\pm$ 1.9 & \textbf{79.1} $\pm$ 0.9 & \textbf{75.3} $\pm$ 1.1 \\
\cline{2-8}
 & ERM-Aug~\cite{gulrajani2020search} & 76.6 $\pm$ 1.5 & 68.0 $\pm$ 2.0 & 70.6 $\pm$ 2.2 & 53.8 $\pm$ 1.6 & 72.2 $\pm$ 2.4 & 68.2 $\pm$ 2.0 \\
 & Macenko-Aug~\cite{macenko2009method} & 70.9 $\pm$ 3.1 & 65.8 $\pm$ 2.7 & 72.4 $\pm$ 1.9 & 54.3 $\pm$ 2.4 & 73.5 $\pm$ 2.0 & 67.4 $\pm$ 2.4 \\
 & RSC-Aug~\cite{huang2020self} & 73.3 $\pm$ 2.8 & 68.5 $\pm$ 2.5 & 70.6 $\pm$ 1.8 & 55.3 $\pm$ 2.0 & 74.7 $\pm$ 2.0 & 68.5 $\pm$ 2.2 \\
 & L2D-Aug~\cite{wang2021learning} & 77.0 $\pm$ 0.5 & 72.0 $\pm$ 0.6 & 67.1 $\pm$ 1.0 & 61.7 $\pm$ 2.9 & 76.4 $\pm$ 0.7 & 70.8 $\pm$ 1.1 \\
 & SFL-Aug~\cite{tomar2024nuclear} & \textbf{77.5} $\pm$ 0.7 & \textbf{76.5} $\pm$ 3.5 & \textbf{75.9} $\pm$ 1.0 & \textbf{74.6} $\pm$ 0.8 & 76.4 $\pm$ 0.9 & \textbf{76.2} $\pm$ 1.4 \\
 & \textbf{Ours-Aug} & 75.6 $\pm$ 0.6 & 75.5 $\pm$ 1.7 & 74.5 $\pm$ 0.9 & 71.1 $\pm$ 1.3 & \textbf{79.7} $\pm$ 0.8 & 75.3 $\pm$ 1.0 \\
\hline
\end{tabular}}
\label{tab:combined_organs_part2}
\end{table*}

\subsection{Robustness Analysis}
To assess the robustness of our model, we conducted performance tests under both natural image distortions and adversarial perturbations.
\subsubsection{Image Corruption Robustness}
\begin{figure*}[t]
    \centering
    \includegraphics[width=0.99999\textwidth]{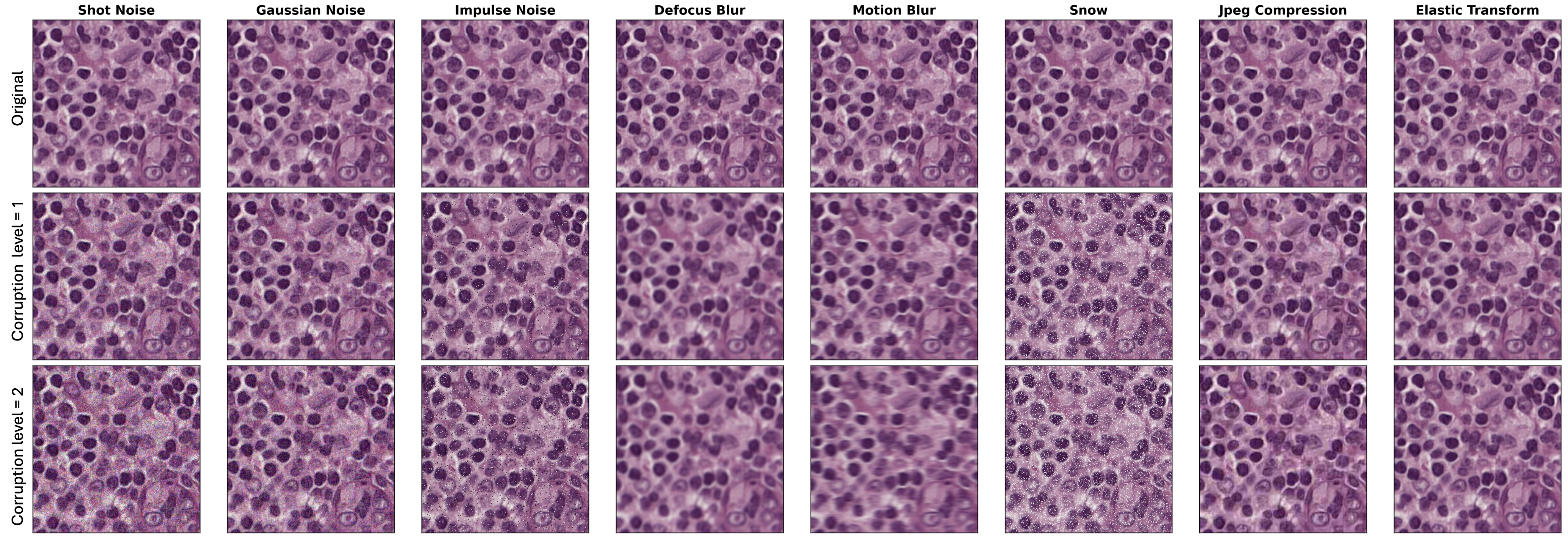}
    \caption{The first row shows the original (uncorrupted) histopathological patches. The second row displays the same patches with a corruption level of 1, and the third row shows patches with a corruption level of 2, as described in \cite{hendrycks2019benchmarking}. Each column corresponds to a specific type of corruption (Best viewed in color).}
    \label{fig:noisy_images_1}
\end{figure*}

\begin{figure*}[t]
    \centering
    \includegraphics[width=0.99999\textwidth]{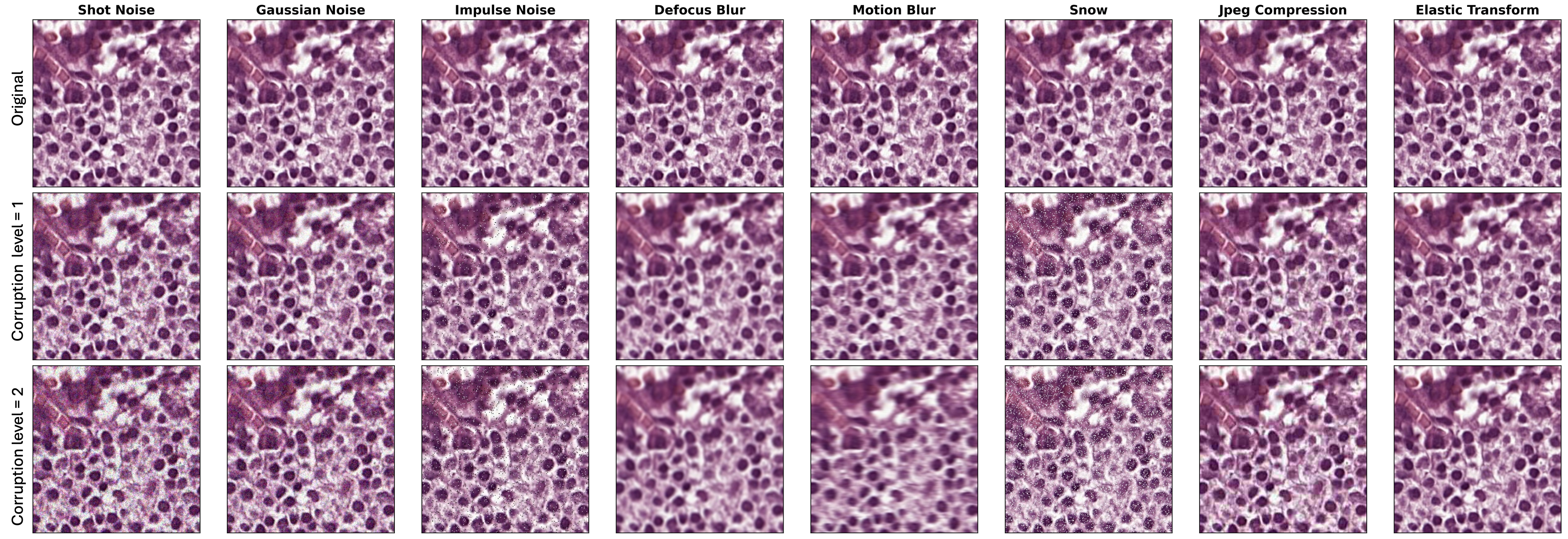}
    \caption{The first row shows the original (uncorrupted) histopathological patches. The second row displays the same patches with a corruption level of 1, and the third row shows patches with a corruption level of 2, as described in \cite{hendrycks2019benchmarking}. Each column corresponds to a specific type of corruption (Best viewed in color).}
    \label{fig:noisy_images_2}
\end{figure*}

\begin{figure*}[ht]
    \centering
    \includegraphics[width=0.99999\textwidth]{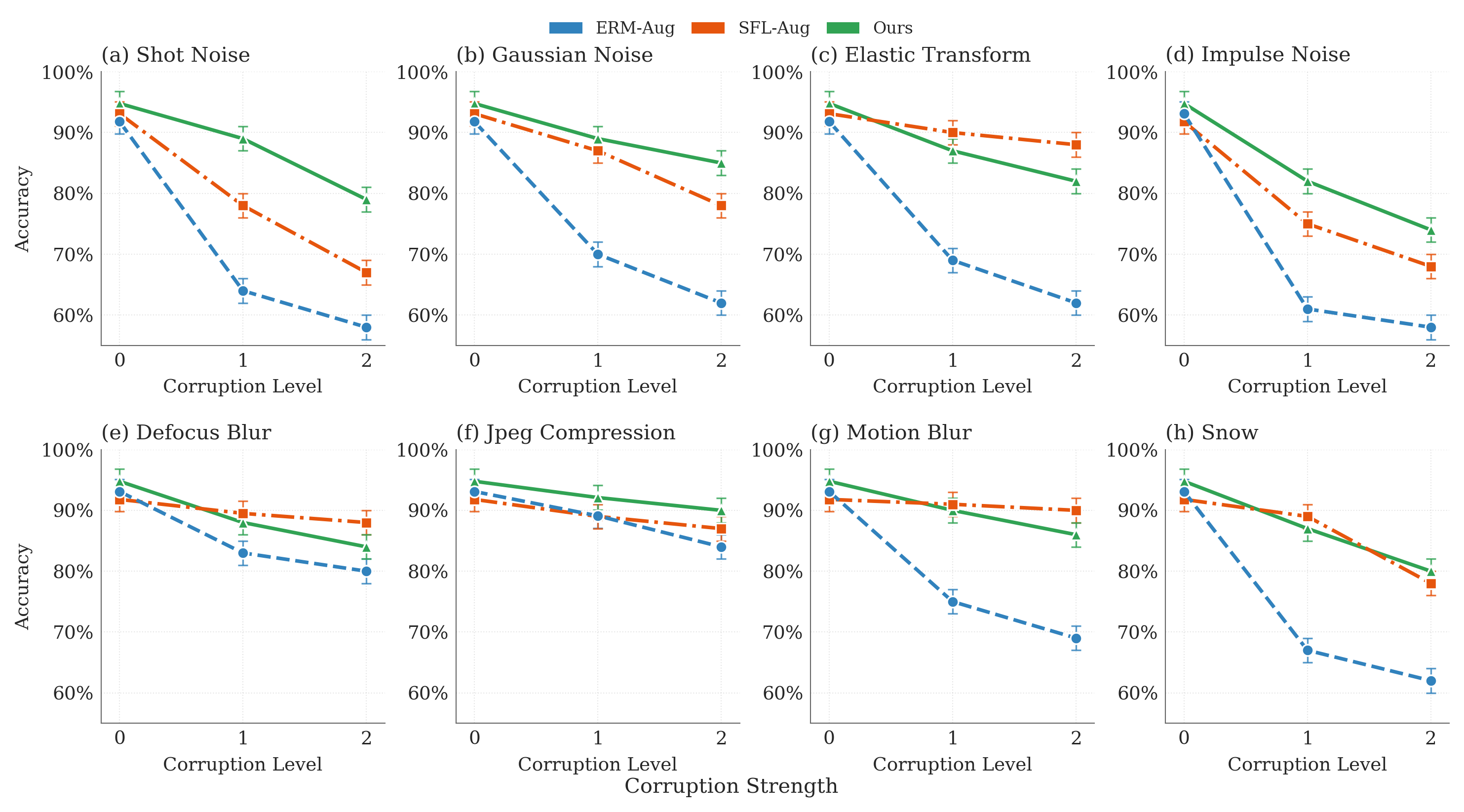}
    \caption{Robustness to common image corruptions on the CAMELYON17 dataset, evaluated using the corruption protocol from \cite{hendrycks2019benchmarking}. Our proposed MorphGen method is compared with baseline approaches across various corruption types (e.g., noise, blur, compression, weather effects) and severity levels. MorphGen consistently maintains higher accuracy, demonstrating strong resilience to diverse perturbations that mimic real-world histopathology artifacts.}
    \label{fig:robustness_to_noise}
\end{figure*}

We investigated eight common natural distortions that typically occur in histopathology imaging during domain shifts. Figures \ref{fig:noisy_images_1} and \ref{fig:noisy_images_2} illustrate example patches from two whole-slide images (WSIs) at two levels of distortion severity: level 0 (no distortion) and level 3 (severe distortion). Figure \ref{fig:robustness_to_noise} provides a comparative analysis of our model against baseline methods under these eight distortions at severity level 3. As shown in Figure \ref{fig:robustness_to_noise}, our morphology-guided representation learning, coupled with a convergence to a flatter optimum, enabled the model to sustain robust performance across five out of the eight distortion types. In contrast, the baseline models exhibited a decline in performance as corruption levels increased, underscoring their sensitivity to domain-specific artifacts and dependence on domain-specialized features. Our approach, which highlights domain-invariant morphological characteristics, demonstrates superior consistency, making it particularly well-suited for deployment in various real-world clinical settings.

\begin{figure*}[ht]
    \centering
    \includegraphics[width=0.925\textwidth]{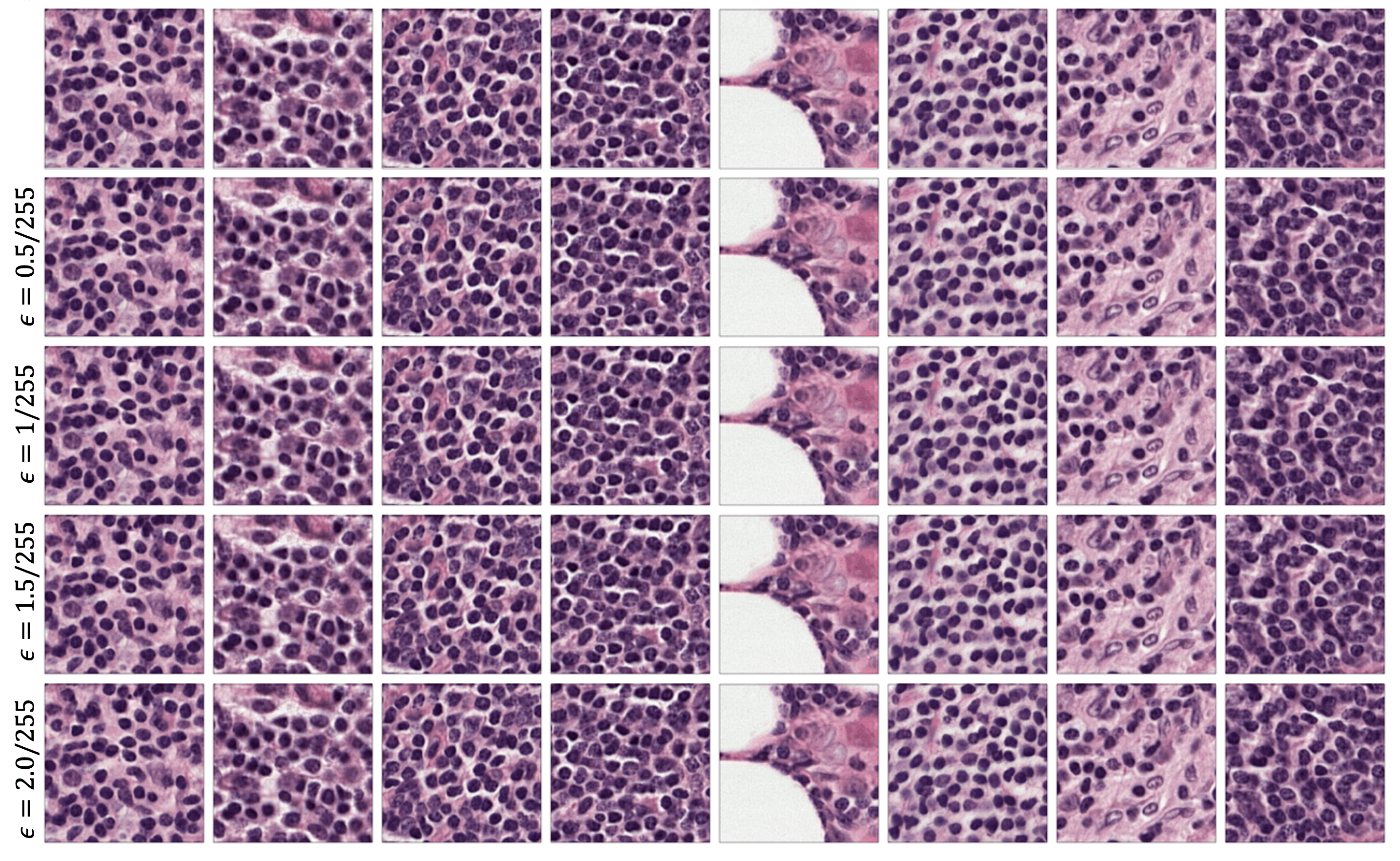}
    \caption{The first row shows the original histopathological images. The second, third, and fourth rows display adversarially perturbed images generated using Projected Gradient Descent (PGD) with increasing epsilon values (perturbation strengths) of 0.5/255, 1/255, 1.5/255, and 2.0/255, respectively.}
    \label{fig:robustness_noise}
\end{figure*} 

\begin{figure*}[ht]
    \centering
    \includegraphics[width=0.5\textwidth]{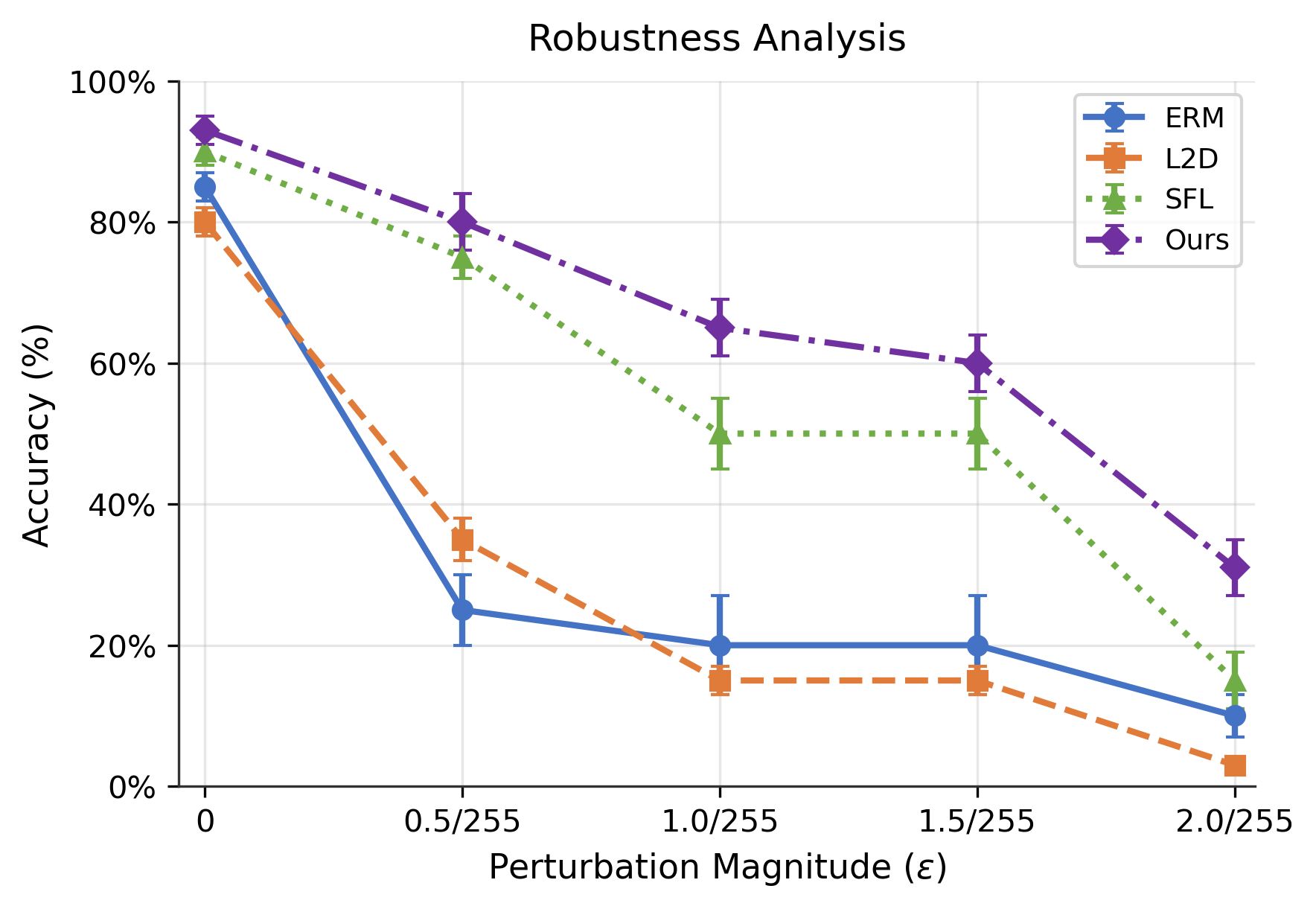}
    \caption{Robustness to image perturbations (e.g., staining artifacts) as described in \cite{hendrycks2019benchmarking, tomar2024nuclear}. Results are averaged over the validation subsets of each center in the CAMELYON17 dataset. Across all four perturbation magnitudes, our proposed MorphGen method consistently outperforms baseline approaches, maintaining substantially higher accuracy under increasing noise severity.}
    \label{fig:robustness_result}
\end{figure*} 

\subsubsection{Adversarial Attack Robustness}
To evaluate the robustness of the learned representations against adversarial perturbations, we follow the protocol outlined in \cite{khan2022adversarially, khan2024adversarially,adversarial_robustness, khan2023importance, khan2022susceptibility}. Specifically, we employ Projected Gradient Descent (PGD) to create three adversarially perturbed versions of the original histopathology patches, each with incrementally increased epsilon values \cite{madry2017towards}. Higher epsilon values indicate stronger adversarial attacks. Figure \ref{fig:robustness_noise} displays sample histopathological images subjected to varying levels of adversarial perturbation, while Figure \ref{fig:robustness_result} compares the performance of our proposed method with that of baseline models under these conditions. As illustrated in Figure \ref{fig:robustness_result}, our method consistently demonstrates the highest accuracy across all three levels of adversarial corruption. The second-best performance is achieved by SFL, which also employs representation alignment through Euclidean distance; however, it remains sub-optimal compared to our approach. This discrepancy arises from our method's dual focus on morphology-guided contrastive learning and strengthening the model’s capacity to learn biologically relevant and domain-invariant features. Consequently, our model is better equipped to handle real-world clinical domain shifts, artifacts, and perturbations resulting from scanning protocols and imaging device variations.

\subsection{Quantitative Analysis (Interpretability via Attention Map)}
\begin{figure*}[ht]
     \centering
     \includegraphics[width=0.99999\textwidth]{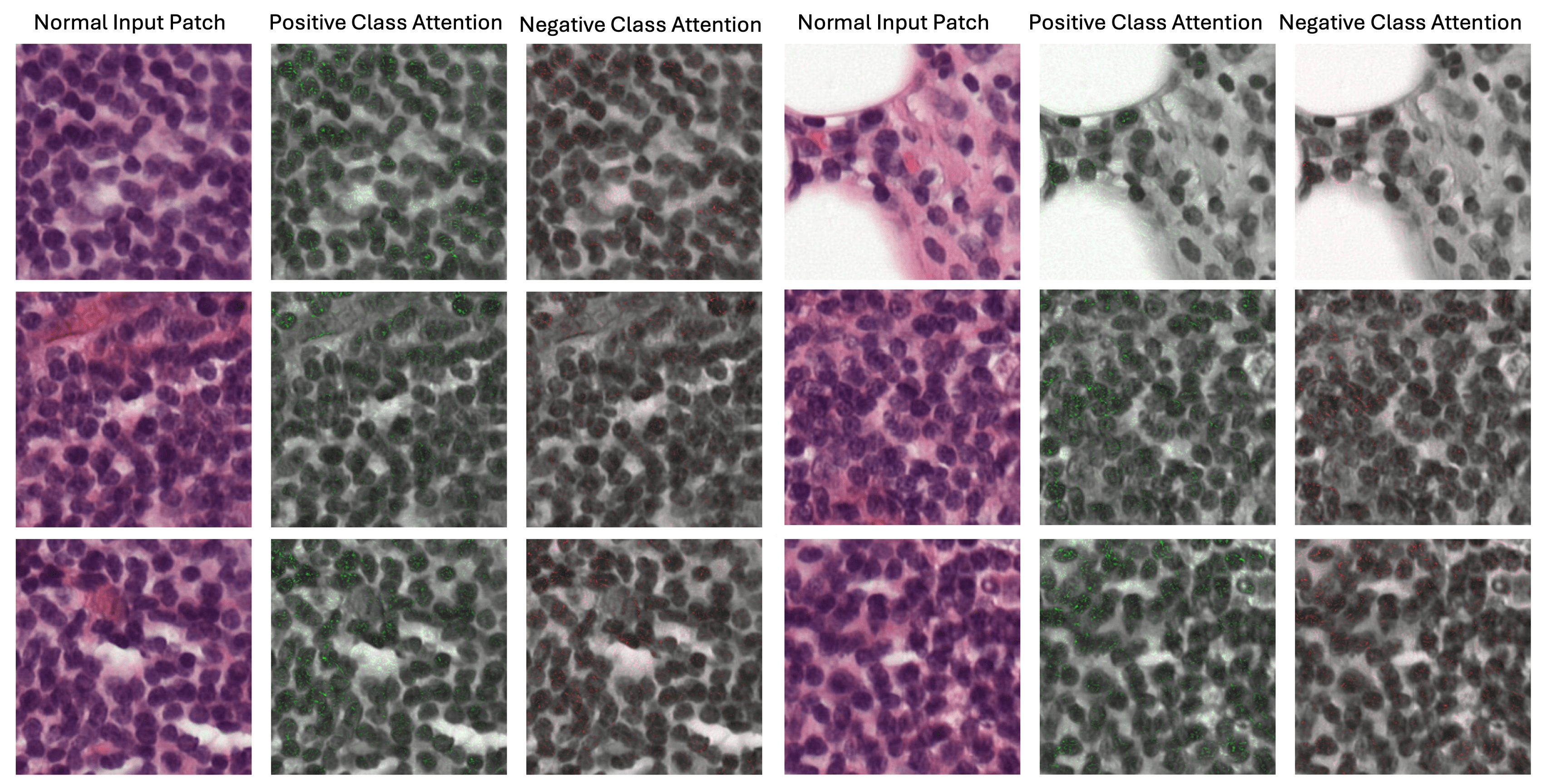}
     \caption{The input patches belong to slide-level negative (no-tumor) whole-slide images (WSIs) from the CAMELYON17 dataset. The first and fourth columns display the original input patches. The second and fifth columns show the model’s attention maps for the positive (tumor) class, while the third and sixth columns illustrate the model's attention maps for the negative (non-tumor) class corresponding to the same inputs.}
     \label{fig:neg_atn_map}
 \end{figure*}

\begin{figure*}[ht]
     \centering
     \includegraphics[width=0.99999\textwidth]{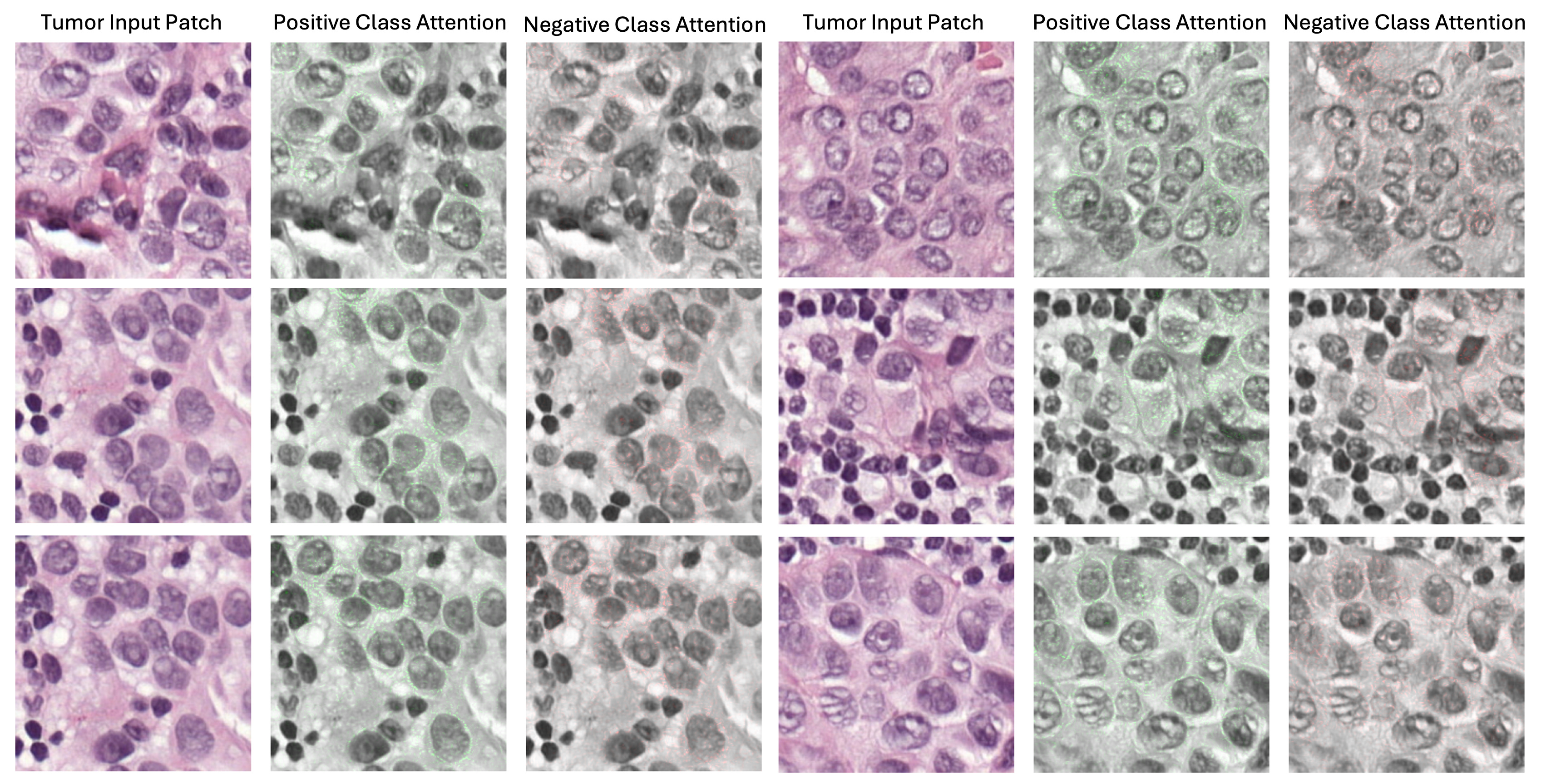}
     \caption{The input patches belong to slide-level positive (with-tumor) whole-slide images (WSIs) from the CAMELYON17 dataset. The first and fourth columns display the original input patches. The second and fifth columns show the model’s attention maps for the positive (tumor) class, while the third and sixth columns illustrate the model's attention maps for the negative (non-tumor) class corresponding to the same inputs.}
     \label{fig:pos_atn_map}
 \end{figure*}

Figures \ref{fig:neg_atn_map} and \ref{fig:pos_atn_map} illustrate the attention maps obtained using the integrated gradients method for normal and tumor tissue patches, respectively \cite{sundararajan2020many}. The model primarily attends to global contextual features for tumor classification, capturing morphological irregularities and the spatial arrangement of heterogeneous cellular populations—hallmarks of malignancy such as disrupted glandular structures, nuclear pleomorphism, and stromal invasion. In contrast, for normal tissue, the model emphasizes localized, fine-grained nuclear features, particularly nuclei and cell centers, reflecting the uniform nuclear morphology and preserved tissue organization typical of healthy samples. These attention patterns align closely with the well-established histopathological criteria used in clinical diagnosis, including nuclear atypia, spatial disorganization, and structural abnormalities, features that exhibit robustness to variations in staining and imaging conditions. This consistency supports our hypothesis that aligning model attention with nuclear morphology and spatial structure facilitates the learning of biologically grounded, domain-invariant representations. Furthermore, the interpretability analysis shows that the learned features not only yield strong classification performance but also mirror diagnostic reasoning strategies employed by expert pathologists. This alignment between model behavior and human expertise enhances the clinical plausibility and transparency of the representations. Collectively, these findings substantiate the central premise of MorphGen: that incorporating morphology-driven constraints into model training fosters interpretable, generalizable, and trustworthy AI systems for digital pathology. The strong correspondence between model attention and expert-defined diagnostic cues underscores the translational potential of our approach and supports its seamless integration into real-world clinical workflows.

In summary, MorphGen demonstrates substantial performance improvements over existing domain generalization and adaptation techniques, supporting our hypothesis that nuclear morphology and spatial organization are robust, domain-invariant features for cancer classification. Across three benchmark datasets—CAMELYON17, BCSS, and OCELOT—MorphGen consistently outperforms state-of-the-art baselines in both out-of-domain and organ-level generalization. Additionally, it exhibits enhanced resilience to natural corruptions (e.g., staining artifacts) and adversarial perturbations. By anchoring representation learning in biologically meaningful cues and enforcing cross-domain alignment in the latent space, MorphGen achieves stain-agnostic, scanner-agnostic, and institution-agnostic generalization—without the need for synthetic augmentation or handcrafted stain normalization. These capabilities underscore its strong potential for deployment in real-world, multi-institutional digital pathology workflows.

\subsection{Clinical Impact}
MorphGen has the potential to significantly advance clinical practice in histopathological cancer diagnosis. By embedding biologically meaningful morphological cues—such as nuclear atypia and tissue architecture—into the training process, MorphGen enhances model interpretability, reliability, and cross-domain generalizability. Unlike many existing models that degrade in performance when exposed to variations in staining protocols or scanner technologies across institutions, MorphGen maintains robust diagnostic accuracy even in unseen domains. This makes it particularly well-suited for deployment in real-world healthcare settings. Such robustness is critically important for under-resourced or rural healthcare systems, where data heterogeneity and limited access to large, annotated datasets often hinder the adoption of AI tools. Clinically, MorphGen can support pathologists by automatically highlighting diagnostically relevant regions, thereby reducing inter-observer variability and enhancing diagnostic throughput—especially valuable in high-volume workflows. Furthermore, its ability to generalize without requiring target domain data eliminates the need for site-specific calibration, lowering technical barriers to implementation. In sum, MorphGen offers a scalable, domain-agnostic solution that can bolster early cancer detection, streamline clinical decision-making, and promote equitable access to high-quality diagnostic care across diverse healthcare environments.

\subsection{Limitations and Future Work}
Although MorphGen exhibits strong out-of-domain generalization and morphological robustness across diverse histopathological datasets, several limitations remain that guide future development. First, the framework requires high-quality nuclear segmentation masks during training. While these annotations facilitate biologically informed learning, this reliance on dense annotations limits scalability in real-world settings, where such masks may be unavailable, inconsistently generated, or prone to interobserver variability. Segmentation errors can also degrade the quality of the representation, adversely affecting downstream performance. Second, the current model is limited to binary classification, restricting its applicability to more complex diagnostic tasks such as tumor subtyping, histological grading, and co-morbidity detection—key requirements for precision oncology. Third, the dual-branch contrastive architecture introduces significant training overhead, with large batch size and memory demands that hinder deployment in resource-constrained environments or at scale. To address these challenges, future work will explore annotation-light strategies such as weak and self-supervised learning, enabling morphological learning without dense segmentation. We will also extend the framework to support multiclass, multilabel, and hierarchical tasks to broaden its clinical relevance. To improve efficiency, we plan to incorporate mixed-precision training, memory-optimized contrastive schemes, and model compression techniques (e.g., pruning, distillation, neural architecture search). Clinical validation in prospective workflows will be essential to evaluate MorphGen’s diagnostic impact, interpretability, and deployment feasibility.

\section{Conclusion}
In this paper, we presented MorphGen, a morphology-guided representation alignment framework designed to address the persistent challenge of single-domain generalization in histopathological cancer classification. By explicitly incorporating biologically grounded, domain-invariant features—such as nuclear morphology, spatial organization, and tissue architecture—MorphGen encourages the learning of robust and generalizable representations that are resilient to variations in staining protocols and imaging conditions. Built upon a supervised contrastive learning paradigm, MorphGen aligns latent representations of original and augmented histopathology images with their corresponding nuclear masks, effectively guiding the model to prioritize diagnostically relevant morphological features. Through comprehensive evaluations across three benchmark datasets and six organ types, we demonstrate that MorphGen consistently surpasses existing baselines in out-of-domain classification accuracy. Interpretability analyses further reveal that the model focuses on meaningful histological cues, reinforcing its alignment with expert pathology practice. Notably, MorphGen exhibits robustness not only to domain shifts but also to common real-world perturbations such as staining artifacts and adversarial attacks—addressing critical limitations of current deep learning models in digital pathology. This domain-agnostic nature reduces reliance on target-specific calibration, significantly lowering deployment barriers and making it well-suited for integration into diverse and resource-constrained clinical environments. Ultimately, MorphGen provides a promising foundation for future work, including the expansion into multi-class and multi-organ cancer classification, the incorporation of broader histological context, and seamless integration with clinical decision support systems, thereby advancing the field toward trustworthy, interpretable, and scalable AI applications in pathology.

\section*{Data Availability}
The code, datasets, and trained models are publicly available at:\newline \href{https://github.com/hikmatkhan/MorphGen/}{https://github.com/hikmatkhan/MorphGen/}.












\bibliography{references}
\bibliographystyle{elsarticle-num}
\end{document}